\documentclass[journal]{IEEEtran}
\usepackage{cite}
\usepackage{graphicx} 
\usepackage{subcaption}
\usepackage{mathptmx} 
\DeclareMathAlphabet{\mathcal}{OMS}{cmsy}{m}{n}
\DeclareSymbolFont{largesymbols}{OMX}{cmex}{m}{n}
\usepackage{amssymb} 
\usepackage{mathtools}
\usepackage[table,xcolor]{xcolor}
\usepackage{threeparttable}
\usepackage{booktabs}
\usepackage[T1]{fontenc}

\usepackage{float}
\usepackage{mathrsfs}
\usepackage{multirow}
\usepackage{makecell}
\usepackage{algorithm}
\usepackage{algorithmic}
\usepackage{enumitem}
\usepackage{amsmath}
\usepackage{amsthm}
\usepackage{hyperref}
\definecolor{custompink}{RGB}{239,118,186} 
\hypersetup{
    colorlinks=true, 
    linkcolor=blue,  
    citecolor=green, 
    filecolor=magenta, 
    urlcolor=custompink     
}

\title{MINER-RRT*: A Hierarchical and Fast Trajectory Planning Framework in 3D Cluttered Environments}

\author{Pengyu Wang$^{1,2}$, \textit{Student Member, IEEE}, Jiawei Tang$^{2}$, \textit{Student Member, IEEE}, Hin Wang Lin$^{2}$, \\ Fan Zhang$^{2}$, Chaoqun Wang$^{4}$, \textit{Member, IEEE}, Jiankun Wang$^{1}$, \textit{Senior Member, IEEE}, \\  Ling Shi$^{2}$, \textit{Fellow, IEEE} and Max Q.-H. Meng$^{1,3}$, \textit{Fellow, IEEE}

\thanks{$^{1}$Pengyu Wang, Jiankun Wang and Max Q.-H. Meng are with Shenzhen Key Laboratory of Robotics Perception and Intelligence and the Department of Electronic and Electrical Engineering, Southern University of Science and Technology, Shenzhen, China. {\tt\small pwangat@connect.ust.hk, wangjk@sustech.edu.cn, max.meng@ieee.org}}%
\thanks{$^{2}$Pengyu Wang, Jiawei Tang, Hin Wang Lin, Fan Zhang and Ling Shi are with the Department of Electronic and Computer Engineering, Hong Kong University of Science and Technology, Hong Kong SAR. {\tt\small \{pwangat, jtangas, hwlinaa, fzhangaw, eesling\}@ust.hk}}%
\thanks{$^{3}$Max Q.-H. Meng is also a Professor Emeritus in the Department of Electronic Engineering at The Chinese University of Hong Kong in Hong Kong and was a Professor in the Department of Electrical and Computer Engineering at the University of Alberta in Canada. {\tt\small max.meng@ieee.org}}%
\thanks{$^{4}$Chaoqun Wang is with the School of Control Science and
Engineering, Shandong University, Shandong, China. {\tt\small chaoqunwang@sdu.edu.cn}}%
}

\begin{document}

\maketitle
\thispagestyle{empty}
\pagestyle{empty}

\begin{abstract}

Trajectory planning for quadrotors in cluttered environments has been challenging in recent years. While many trajectory planning frameworks have been successful, there still exists potential for improvements, particularly in enhancing the speed of generating efficient trajectories. In this paper, we present a novel hierarchical trajectory planning framework to reduce computational time and memory usage called MINER-RRT*, which consists of two main components. First, we propose a sampling-based path planning method boosted by neural networks, where the predicted heuristic region accelerates the convergence of rapidly-exploring random trees. Second, we utilize the optimal conditions derived from the quadrotor's differential flatness properties to construct polynomial trajectories that minimize control effort in multiple stages. Extensive simulation and real-world experimental results demonstrate that, compared to several state-of-the-art (SOTA) approaches, our method can generate high-quality trajectories with better performance in 3D cluttered environments. (Video\footnote{\href{https://youtu.be/fXuuMRX19q0}{https://youtu.be/fXuuMRX19q0}})


\end{abstract}

\def\abstractname{Note to Practitioners}
\begin{abstract}
The motivation is the problem of planning trajectories for quadrotor autonomous flight in 3D cluttered and complex scenarios such as wild forest exploration and subterranean environment search-and-rescue. Sampling-based path planning methods are suitable for dealing with the complexity of the physical environment but are not convenient for computing dynamics and their differentials. Optimization-based trajectory generation methods are appropriate for handling various high-order constraints but rely on high-quality initial path solutions. Therefore, this paper combines the advantages of the two methods to propose a novel trajectory planning framework that can generate high-quality trajectories for quadrotors faster than many previous algorithms. We conduct numerous simulations and real-world experiments to verify that our method can be effectively deployed in real scenarios and empower quadrotors for complex autonomous tasks in the future.
\end{abstract}

\def\abstractname{Index Terms}
\begin{abstract}
Robot trajectory planning, sampling-based algorithm, deep neural network.
\end{abstract}

\section{INTRODUCTION}
\IEEEPARstart{U}{nmanned} Aerial Vehicles (UAVs) have been increasingly utilized in various applications recently, including environmental surveillance~\cite{li2019coverage}, search-and-rescue~\cite{ngo2022uav}, package delivery~\cite{sheng2023unmanned}, and air-ground collaboration~\cite{wang2022quadrotor}. These complex scenarios pose significant challenges for the autonomous flight of UAVs, among which trajectory planning is one of the most critical modules for UAV's autonomy. Therefore, how to quickly and efficiently plan high-quality trajectories has become a great concern for researchers.

\begin{figure}[t!]
    \centering
    \begin{subfigure}{0.48\columnwidth}
        \includegraphics[width=\textwidth]{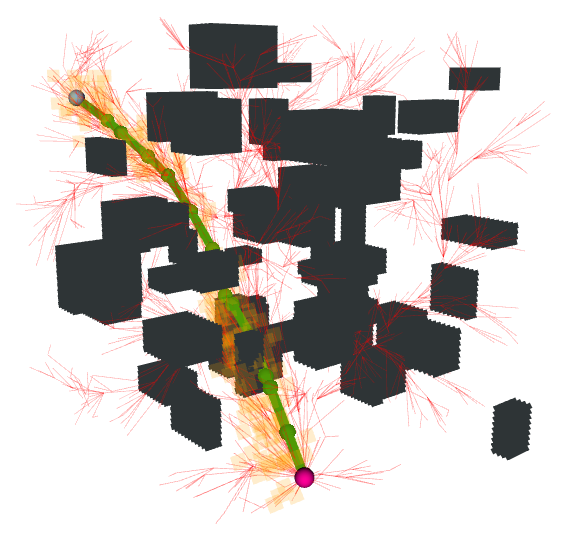}
        \caption{}
        \label{fig:show1}
    \end{subfigure}
    \begin{subfigure}{0.48\columnwidth}
        \includegraphics[width=\textwidth]{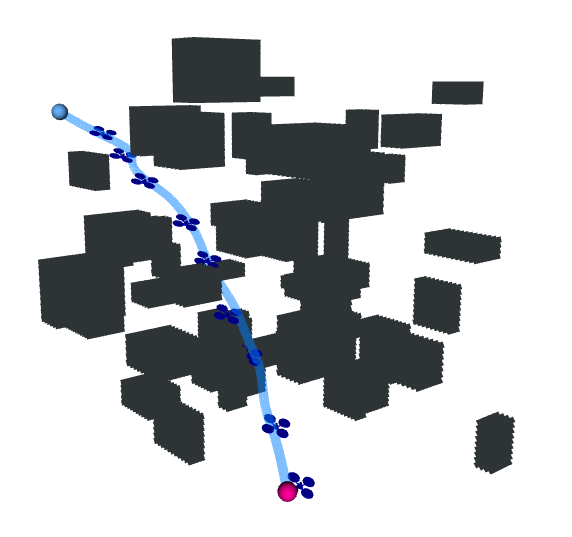}
        \caption{}
        \label{fig:show2}
    \end{subfigure}
    \caption{Illustration of our hierarchical trajectory planning framework MINER-RRT*. (a) Front end: The predicted heuristic region is shown in orange, the edges of the RRT are shown in light red, and the generated initial path is shown in green. (b) Back end: The resulting minimum snap trajectory is shown in light blue.}
    \label{fig:show}
\end{figure}

Three types of path planning methods are commonly used to quickly find initial path solutions in complex topological environments. A classical search-based method such as A*~\cite{hart1968formal} achieves optimality based on the degree of discreteness in the configuration space. This resolution optimality characteristic makes these methods time and space-consuming in high-dimensional areas. The artificial potential field~\cite{khatib1986real} allows the robot to move in the direction of the fastest gradient descent, but it is easy to get stuck in a local minimum. Sampling-based method, such as rapidly exploring random tree (RRT)~\cite{lavalle1998rapidly}, continuously explores and exploits environmental information to seek the global optimal solution. RRT*~\cite{karaman2011sampling}, which is the extension of RRT, achieves asymptotic optimality by incrementally re-constructing search trees in continuous space. 
However, converging to the optimal solution takes a lot of time and memory cost, so many heuristic-enhanced methods have been proposed. Informed RRT*~\cite{gammell2018informed} and Guided RRT*~\cite{scalise2023guild} constrain the sampling region to $L^2$ informed sets and local subsets, thereby speeding up the convergence of the original RRT* method, but the performance of both methods heavily depends on high-quality initial solutions and the design of complex, manually-crafted heuristic regions. Our previous work~\cite{wang2020neural} first proposes an end-to-end neural network-based approach to predict heuristic promising regions for RRT* sampling in 2D and then in 3D space~\cite{wang2021deep}. However, neither of the above two works properly considers the guarantee of the heuristic region connectivity (whether the heuristic region contains at least one path connecting the start to the goal) and safety (whether the heuristic region includes obstacles in the environment), thus leading to a decrease in heuristic region quality and ultimately reducing performance in the path planner. Therefore, in response to the previous issues, we propose a novel neural heuristic-based path-planning front end with safe and connected heuristics to speed up the entire trajectory-planning process.

Researchers further extend RRT*-like algorithms to kinodynamics~\cite{webb2013kinodynamic}~\cite{li2016asymptotically}, but it is inevitable to solve two-point boundary value problems, which are usually non-trivial and time-consuming. Therefore, in this paper, we only use the ability of the RRT* algorithm to deal with complex physical environments. Since the sampling-based methods are inefficient in dealing with the robot's kinodynamics and its differential constraints, researchers use optimization-based methods and exploit high-order information to make the trajectory converge to the optimal solution of specific problem formulations. For quadrotor UAV, its differential flatness~\cite{van1998real}~\cite{mellinger2011minimum} allows the state and input to be written as an algebraic function of the flat output and its derivative, and the quadrotor can directly follow the trajectory obtained in the flat space. Polynomial functions are widely used to parameterize trajectories in flat space due to their characteristics of easy definition of smoothness criteria, easy calculation of derivatives, and easy decoupling in each dimension of independent flat output. Mellinger and Kumar~\cite{mellinger2011minimum} pioneer to formulate a quadratic programming (QP) problem for minimizing the fourth-order time derivative (snap) of the polynomial splines. To avoid numerical issues that arise as the number of trajectory segments increases, Bry \textit{et al.}~\cite{bry2015aggressive} and Oleynikova \textit{et al.}~\cite{oleynikova2020open} directly optimize the end-derivatives of segments present within the spline and provide an unconstrained closed-form solution to the problem above. However, the singularity of inverting matrices is not well discussed. Wang \textit{et al.}~\cite{wang2022geometrically} further provides the necessary and sufficient optimality condition for this type of problem to directly construct trajectories, and we utilize this condition to further accelerate the whole planning process. However, all polynomial trajectory optimization methods above rely on constructing configuration space and high-quality initial trajectory. 
When the working scene is cluttered and complex in 3D, constructing the configuration space consumes many resources, making it more difficult to generate feasible initial trajectories.

Therefore, we first utilize the sampling-based RRT* algorithm to quickly find the initial path in cluttered environments and avoid the precise construction of configuration spaces, where this process is called frond-end and is shown in Fig.~\ref{fig:show1}. Then, we assign the initial trajectory segments a reasonable time allocation by a designed trapezoidal scheme. Finally, we combine the polynomial trajectory optimization method and take the initial trajectory, time allocation, and different constraints as input to generate the final trajectory, where this process is called back-end and is shown in Fig.~\ref{fig:show2}. In summary, we propose a novel and unified trajectory planning framework to \textbf{MIN}imize control effort, including a n\textbf{E}u\textbf{R}al-boosted \textbf{RRT*} front-end and an optimal condition-based back-end, which is called \textbf{MINER-RRT*}. The main contributions of this paper are multifold:

\begin{enumerate}[label=\arabic*)]
  \item We propose a hierarchical and fast trajectory planning framework in 3D cluttered environments, which can generate trajectories for autonomous flight while satisfying the corresponding constraints.
  \item We propose an efficient learning enhanced sampling-based method with specially designed loss functions to find initial path solutions in 3D topological environments quickly.
  \item We exploit optimal condition-based trajectory optimization methods as our back end to further accelerate generating final trajectories that minimize control efforts.
\end{enumerate}

The remainder of this paper is organized as follows. We first do a literature review on sampling-based path planning and optimization-based polynomial trajectory generation methods in further detail in Section II. In Section III, we give some preliminaries and formulations of the trajectory planning framework. In Section IV, we elaborate on the details of the proposed hierarchical planning method. Moreover, we conduct extensive comparative experiments and data analysis in Section V. Finally, we draw a conclusion and discuss our future work in Section VI.

\section{RELATED WORK}

\subsection{Sampling-based Path Planning}

Initially, feedback controllers, such as~\cite{tedrake2010lqr}, achieve motion planning by constructing a set of controllers to reach the target point; however, these methods are unable to perform obstacle avoidance in unknown environments. In contrast, the core of sampling-based RRT is to build a tree of feasible vertices and edges incrementally to obtain the optimal solution. Compared with RRT, RRT* re-selects the parent nodes and rewires them according to the optimal cost-from-start values, which can guarantee asymptotic optimality. Arslan and Tsiotras~\cite{arslan2013use} further propose RRT\# to address the issues of over-exploitation and under-exploitation found in RRT*. Qureshi and Ayaz~\cite{qureshi2015intelligent} propose IB-RRT*, which adopts a bidirectional tree strategy with a smart sampling heuristic for fast convergence. To avoid the non-trivial two-point boundary value problem, authors in~\cite{nayak2022bidirectional} present a probabilistically complete method that uses a backward tree as heuristics to guide the growth of the forward tree. RRTX is proposed in~\cite{otte2016rrtx} to achieve fast re-planning, thereby accommodating dynamic obstacles in the environment. Similarly, in~\cite{jaffar2022pip}, the authors propose a real-time feedback re-planning algorithm using invariant sets "funnels". Recent works further improve the sampling strategy by designing different heuristics. Informed RRT* employs a direct sampling technique within the informed set once a feasible path is discovered. 
Batch Informed Trees (BIT*)~\cite{gammell2020batch} is an informed and anytime sampling-based planner. It alternates between sampling and heuristic techniques to estimate and explore the problem domain. The authors in~\cite{strub2022adaptively} introduce Adaptively Informed Trees (AIT*) and Effort Informed Trees (EIT*) based on BIT*. These two asymptotically optimal methods enhance planning by concurrently calculating and utilizing problem-specific heuristics. Recently, Scalise \textit{et al.}~\cite{scalise2023guild} develop a local densification technique to enhance the efficiency of the informed set by employing subsets defined by beacons.

In recent years, learning-based approaches have been integrated into path-planning tasks and are generally divided into two categories. The first category involves the use of reinforcement learning (RL)-based methods to enhance the performance of planning. Chiang \textit{et al.} proposed the RL-RRT~\cite{chiang2019rl} method, which uses an RL approach based on deep deterministic policy gradients (DDPG) combined with upper-layer sampling-based planning methods to achieve kinodynamic planning. In~\cite{kontoudis2019kinodynamic}, the authors leverage an actor-critic (AC) policy neural network to solve the boundary value problem online. To tackle completely unknown system dynamics, Xu \textit{et al.}~\cite{xu2023online} present an intermittent Q-learning method as a controller between waypoints and combine it with a sampling-based planner to achieve kinodynamic planning. Another category of methods uses deep supervised learning to employ representations of state space and change the sampling strategy of RRT*-based methods to accelerate the planning procedure. Kuo \textit{et al.}~\cite{kuo2018deep} combine a sequence model with a sampling-based planner to affect the next move and state of the planner through observation. In~\cite{allen2019real}, the authors use learning from demonstrations to estimate the reachable sets for specific states, which is ultimately combined with kinodynamic RRT* to significantly reduce computation time. Wang \textit{et al.}~\cite{wang2020neural} propose an end-to-end pixel-based path planning framework based on Convolutional Neural Network (CNN) and predict the promising region where the optimal path probably exists without preprocessing steps. The implementation of a Generative Adversarial Network (GAN) can also be employed to accomplish nonuniform sampling in RRT*-like algorithms~\cite{ma2021conditional}~\cite{li2021efficient}~\cite{zhang2021generative}, but the effectiveness in 3D cluttered environments is unknown. Researchers then
extend~\cite{wang2020neural} to 3D~\cite{wang2021deep} and investigate different deep learning models' performance on heuristic region generation. However, there are insufficient simultaneous constraints on both the safety and connectivity of the generated heuristic regions.

\subsection{Optimization-based Polynomial Trajectory Generation}

For general dynamical systems, many existing solvers can handle trajectory optimization problems and obtain high-quality solutions, such as GPOPS-II~\cite{patterson2014gpops} and ALTRO~\cite{howell2019altro}. However, mobile robots typically involve constraints that are non-smooth and difficult to describe explicitly. Trajectory planning for differentially flat quadrotors can be transformed into an optimization problem of low-dimensional output in flat space. Mellinger and Kumar~\cite{mellinger2011minimum} use fixed-duration polynomial splines to characterize the flat trajectories. 
To obtain the gradient of the cost function for time allocation, the perturbation problem of quadratic programming needs to be solved multiple times. 
Bry \textit{et al.}~\cite{bry2015aggressive} then give a closed-form solution to this quadratic program when there are only higher-order continuity constraints and waypoint constraints. Trajectory segments generated in~\cite{campos2017hybrid} have constant acceleration. The authors derive the speed limit and the maximum difference between the trajectory and the path and formulate it into a convex optimization problem to avoid the iterative procedure. Oleynikova \textit{et al.}~\cite{oleynikova2020open} adopt the method of optimizing all end derivatives directly within the trajectory segment. At the same time, they integrate collision-checking as soft costs into the local continuous optimization. To speed up trajectory generation, Burke \textit{et al.}~\cite{burke2020generating} introduces a linear-complexity scheme in the number of segments to address primal-dual variables of the QP problem. Wang \textit{et al.}~\cite{wang2022geometrically} give the optimal condition of integral chain systems in multi-stage unconstrained minimum control problems. This condition can generate a unique and smooth trajectory for a given group of time and waypoint parameters without computing cost function gradients.

\section{PRELIMINARIES}

\subsection{Optimal Path Planning}

Let $\mathcal{X} \subseteq \mathbb{R}^n$ represents the state configuration space, $\mathcal{X} _{\mathrm{obs}}\subset \mathcal{X}$ denotes the space occupied by obstacles, and $\mathcal{X} _{\mathrm{free}}=closure\left( \mathcal{X} \backslash\mathcal{X} _{\mathrm{obs}} \right)$ corresponds to the unoccupied space. The initial condition and goal condition are represented as $x_{\mathrm{init}}\in \mathcal{X} _{\mathrm{free}}$ and $x_{\mathrm{goal}}\in \mathcal{X} _{\mathrm{goal}}=\left\{ x\in \mathcal{X} |\left\| x-x_{\mathrm{goal}} \right\| <radius \right\}$, respectively~\cite{karaman2011sampling}. 

\textit{Feasible Path Planning Problem}: Consider the triplet $\left( \mathcal{X} _{\mathrm{free}}, x_{\mathrm{init}}, \mathcal{X} _{\mathrm{goal}} \right)$, find a feasible path comprised of collision-free states $\varsigma \left( t \right) :\left[ 0,T \right] \rightarrow \mathcal{X} _{\mathrm{free}}$ such that $\varsigma \left( 0 \right) =x_{\mathrm{init}}$, and $\varsigma \left( T \right) \in \mathcal{X} _{\mathrm{goal}}$.

Let $\varSigma$ denote the set of all feasible paths, and $c\left( \varsigma \right)$ is the cost function for path evaluation. 

\textit{Optimal Path Planning Problem}:
\begin{equation}
\begin{aligned}
    & c\left( \varsigma ^* \right) =\underset{\varsigma \in \varSigma}{\arg\min} \, c\left( \varsigma \right) \\
    & s.t. \varsigma \left( 0 \right) =x_{\mathrm{init}}, \varsigma \left( T \right) \in \mathcal{X} _{\mathrm{goal}}, \varsigma \left( t \right) \in \mathcal{X} _{\mathrm{free}}, \forall t\in \left[ 0,T \right].
\end{aligned}
\end{equation}

\begin{figure*}[t]
    \centering
    \includegraphics[width=\textwidth]{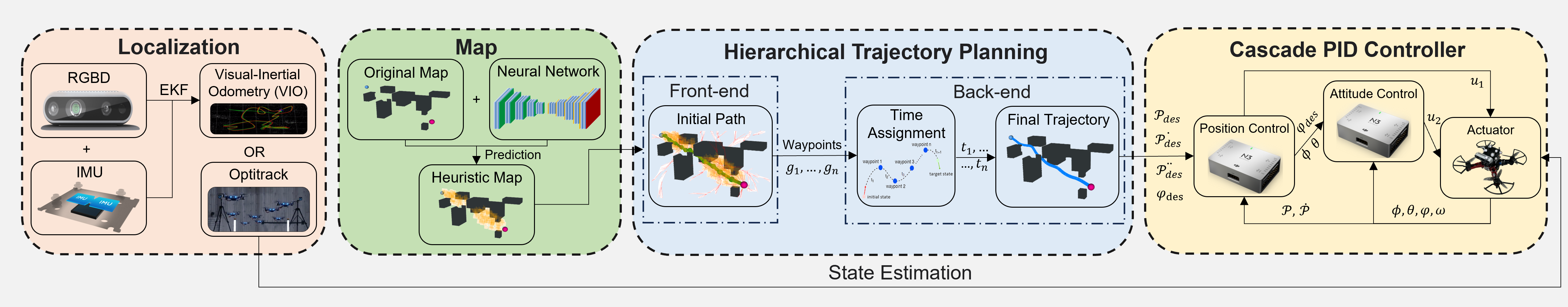}
    \caption{Architecture of our system. The localization module and controller are on both sides of the figure (Sec.~\ref{sec:overview}). The front end of our framework is the initial path finding guided by the heuristic map (Sec.~\ref{sec:Frond-end}). The back end of our framework is the optimal trajectory generation (Sec.~\ref{sec:Back-end}).}
    \label{fig:framework}
\end{figure*}

\subsection{Differential Flatness}
Consider the following nonlinear system:
\begin{equation}
    \dot{x}=f\left( x,u \right),
\end{equation}
where system state $x\in \mathbb{R}^n$ and control input $u\in \mathbb{R}^m$. A differentially flat dynamic system~\cite{van1998real} exhibits the property of having a set of flat outputs $\gamma\in \mathbb{R}^m$, which are uniquely determined by the finite derivatives mapping $\zeta$ of the state $x$ and control $u$. At the same time, $x$ and $u$ can also be parameterized by the flat outputs and their finite derivatives:
\begin{equation}
\begin{aligned}
    & \gamma=\zeta \left( x,u,\dot{u},\cdots ,u^{\left( s \right)} \right), \\
    & x=\varPhi _x\left( \gamma,\dot{\gamma},\cdots ,\gamma^{\left( s-1 \right)} \right), \\
    & u=\varPhi _u\left( \gamma,\dot{\gamma},\cdots ,\gamma^{\left( s \right)} \right),
\end{aligned}
\end{equation}
where the flat mapping $\varPhi _x:\left( \mathbb{R} ^m \right) ^s\mapsto \mathbb{R} ^n
$, $\varPhi _u:\left( \mathbb{R} ^m \right) ^{s+1}\mapsto \mathbb{R} ^m$, and $s$ is the order of the integral chain below. To simplify notation, we put the flat outputs and their derivatives into a flat flag $\gamma^{\left[ s-1 \right]}\in \mathbb{R}^{m\times s}$:
\begin{equation}
    \gamma^{\left[ s-1 \right]}=\left( \gamma^T,\dot{\gamma}^T,\cdots ,\gamma^{\left( s-1 \right) ^T} \right) ^T.
\end{equation}

The flat output of quadrotors highly coincides with the configuration space considered in general motion planning. For various underactuated multi-rotors, the flat output generally has a similar form:
\begin{equation}
    \gamma=\left( p_x,p_y,p_z,\psi \right) ^T,
\end{equation}
where $\left( p_x,p_y,p_z \right) ^T$ is the displacement of the UAV's center of gravity, and $\psi$ is the yaw angle. Based on this, we can directly optimize the trajectory $\gamma \left( t \right) $ in the flat output space and obtain $x\left( t \right)$ and $u\left( t \right)$ through the flat mapping.

\subsection{Full State Trajectory Optimization}

For differentially flat quadrotors, the full-state trajectory optimization that satisfies the boundary-intermediate value problem (BIVP) condition can be formulated as follows. 

\textit{$M$-Segment Unconstrained Control Effort Minimization Problem}~\cite{wang2022geometrically}:

\begin{subequations}
    \begin{align}
        & \underset{\gamma\left( t \right)}{\min}\int_{t_0}^{t_M}{\iota \left( t \right)  ^TQ\iota \left( t \right)  dt} \label{eqn:minco1} \\
        & s.t. \quad \gamma^s\left( t \right) =\iota \left( t \right) , \,\forall t\in \left[ t_0, t_M \right] \label{eqn:minco2},  \\
        & \quad \quad \gamma^{\left[ s-1 \right]}\left( t_0 \right) =\overline{\gamma}_0, \,\gamma^{\left[ s-1 \right]}\left( t_M \right) =\overline{\gamma}_f \label{eqn:minco3}, \\
        & \quad \quad \gamma^{\left[ d_i-1 \right]}\left( t_i \right) =\overline{\gamma}_i, \,1\leqslant i<M \label{eqn:minco4}, \\
        & \quad \quad t_{i-1}<t_i, \,1\leqslant i<M \label{eqn:minco5},
    \end{align}
\end{subequations}
where quadratic form of control effort $\iota \left( t \right) $ is used to achieve trajectory smoothness, time interval $\left[ t_0, t_M \right] $ is divided into $M$ segments by $M + 1$ fixed time stamps, $Q\in \mathbb{R} ^{m\times m}$ is a positive semi-definite matrix, and $\overline{\gamma}_0, \overline{\gamma}_f\in \mathbb{R}^{m\times s}$ are the boundary conditions of the entire trajectory. The intermediate condition $\overline{\gamma}_i\in \mathbb{R}^{m\times d_i}$ must satisfy $d_i\leqslant s$, which specifies the value of the trajectory at time $t_i$ up to the $d_i$-order derivative $\gamma\left( t_i \right) , \dot{\gamma}\left( t_i \right) ,...,\gamma^{\left( d_i-1 \right)}\left( t_i \right) $. 

\section{METHODOLOGY}

\subsection{System Overview} 
\label{sec:overview}

The architecture of our system framework is shown in Fig.~\ref{fig:framework}. Specifically, our framework contains four parts, namely the localization module, the hierarchical trajectory planning module and the controller module. For the localization module, we use the extended Kalman filter to fuse information from binocular depth images and IMU to form visual-inertial odometry (or simply use ground truth pose information directly from the motion capture system). The hierarchical trajectory planning module, which is the main contribution of this paper and will be detailed in Sec.~\ref{sec:Frond-end} - Sec.~\ref{sec:Back-end}, contains a front-end learning-based path-finding algorithm and a back-end trajectory generation algorithm. For the controller module inside the flight controller, we adopt an \( SE(3) \) cascade PID strategy as our control strategy. This strategy includes outer loop position control and inner loop attitude control. For the hardware setting, the quadrotor UAV used in this paper mainly includes a DJI N3 Flight Controller\footnote{https://www.dji.com/hk-en/n3}, a DJI Manifold-2G Computer\footnote{https://www.dji.com/hk-en/manifold-2} and an Intel Realsense D435i Camera\footnote{https://www.intelrealsense.com/depth-camera-d435i}.

\subsection{Trajectory Planning Frond-end: Fast Initial Path Finding}
\label{sec:Frond-end}

\subsubsection{3D-sGAN-RRT*}

As shown in Section I, traditional sampling-based path planning methods are usually slow to find the initial solution, and learning-based methods can be further improved when applied to 3D cluttered environments. Therefore, in this paper, we propose a 3D conditional generative adversarial network combined with specially designed loss functions and an improved RRT* algorithm to increase the front end's convergence speed (denoted by 3D-sGAN-RRT*).

\begin{figure*}[htb]
    \centering
    \begin{subfigure}{0.2\textwidth}
        \includegraphics[width=\textwidth]{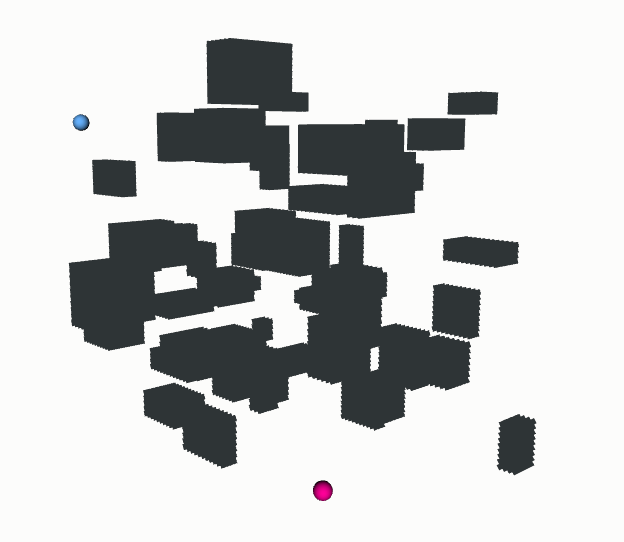}
        \caption{Original map.}
        \label{fig:map1-1}
    \end{subfigure}%
    \begin{subfigure}{0.2\textwidth}
        \includegraphics[width=\textwidth]{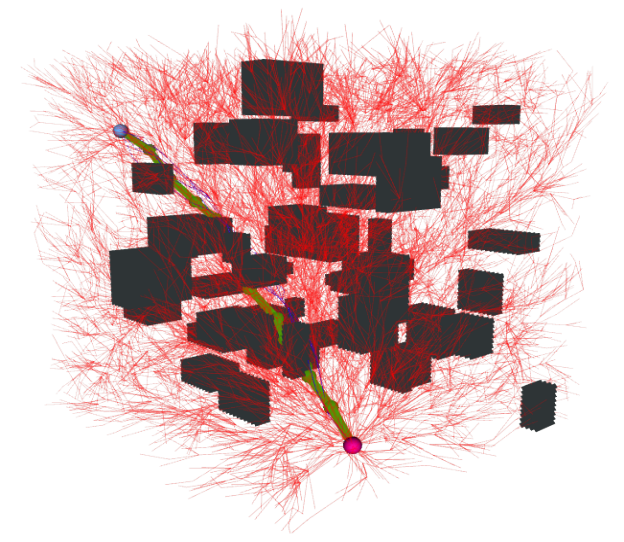}
        \caption{RRT*'s front-end path.}
        \label{fig:map1-2}
    \end{subfigure}%
    \begin{subfigure}{0.2\textwidth}
        \includegraphics[width=\textwidth]{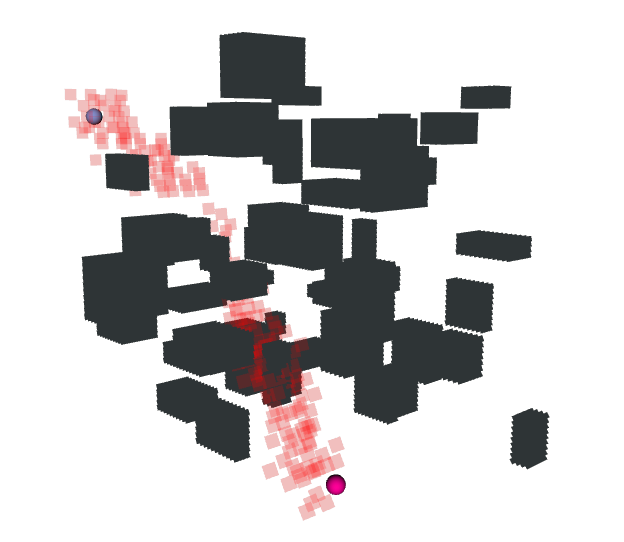}
        \caption{Heuristic region.}
        \label{fig:map1-3}
    \end{subfigure}%
    \begin{subfigure}{0.2\textwidth}
        \includegraphics[width=\textwidth]{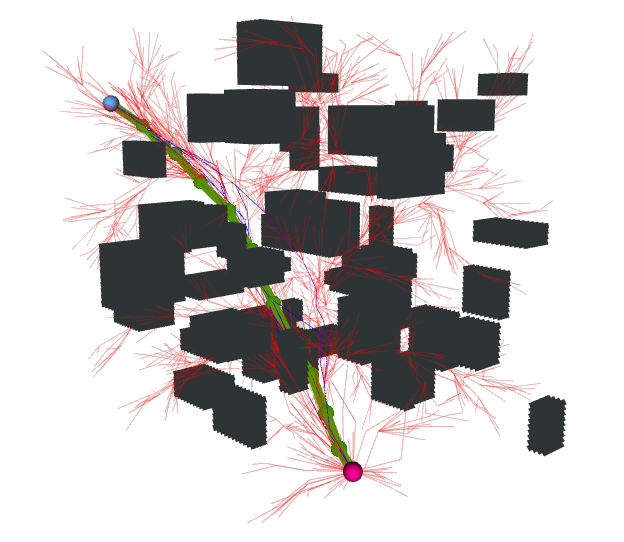}
        \caption{Our front-end path.}
        \label{fig:map1-4}
    \end{subfigure}%
    \begin{subfigure}{0.2\textwidth}
        \includegraphics[width=\textwidth]{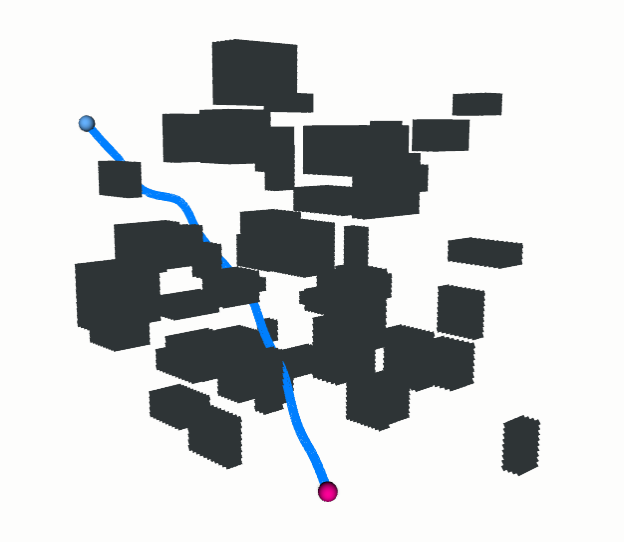}
        \caption{Our final trajectory.}
        \label{fig:map1-5}
    \end{subfigure}
    \caption{The visualization of trajectory planning process: original map, RRT*'s front-end path, heuristic region, our front-end path, our final trajectory (Simulation map1).}
    \label{fig:Map1_vis}
\end{figure*}

\begin{figure*}[htb]
    \centering
    \begin{subfigure}{0.2\textwidth}
        \includegraphics[width=\textwidth]{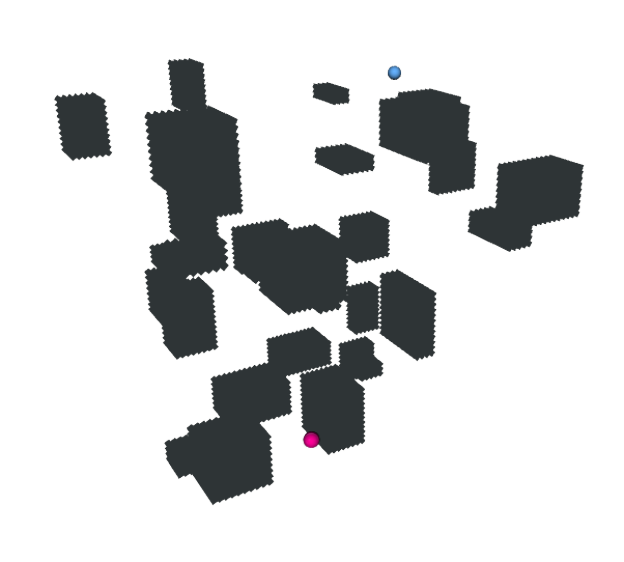}
        \caption{Original map.}
        \label{fig:map2-1}
    \end{subfigure}%
    \begin{subfigure}{0.2\textwidth}
        \includegraphics[width=\textwidth]{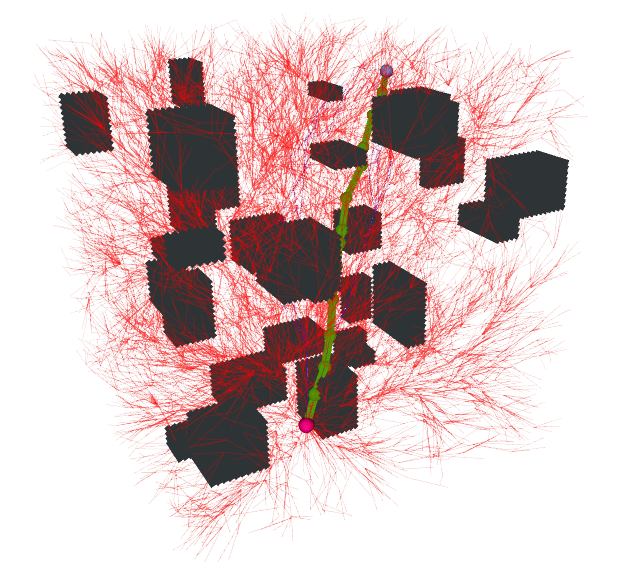}
        \caption{RRT*'s front-end path.}
        \label{fig:map2-2}
    \end{subfigure}%
    \begin{subfigure}{0.2\textwidth}
        \includegraphics[width=\textwidth]{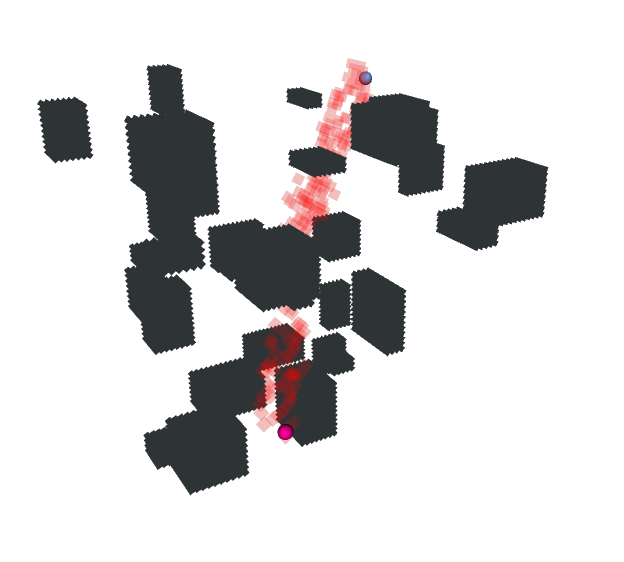}
        \caption{Heuristic region.}
        \label{fig:map2-3}
    \end{subfigure}%
    \begin{subfigure}{0.2\textwidth}
        \includegraphics[width=\textwidth]{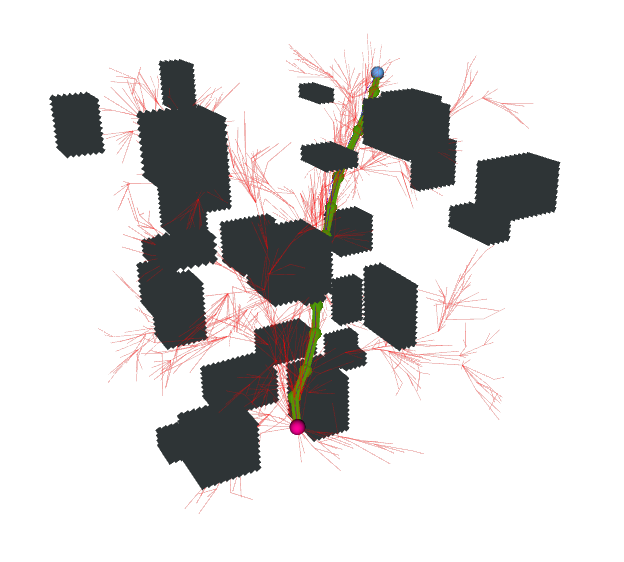}
        \caption{Our front-end path.}
        \label{fig:map2-4}
    \end{subfigure}%
    \begin{subfigure}{0.2\textwidth}
        \includegraphics[width=\textwidth]{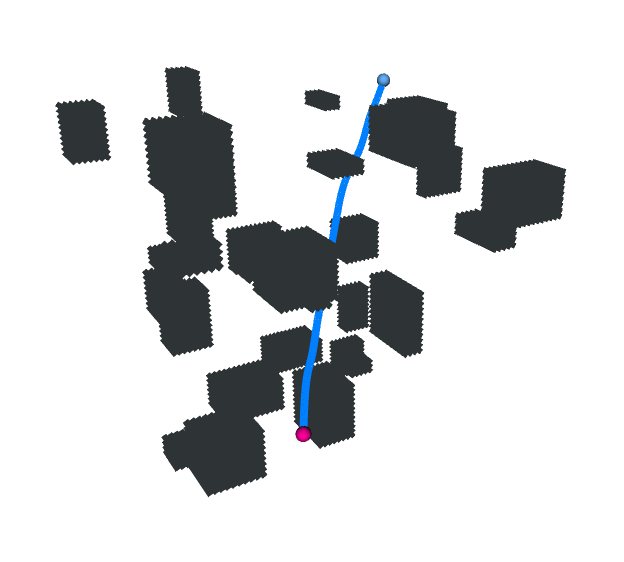}
        \caption{Our final trajectory.}
        \label{fig:map2-5}
    \end{subfigure}
    \caption{The visualization of trajectory planning process: original map, RRT*'s front-end path, heuristic region, our front-end path, our final trajectory (Simulation map2).}
    \label{fig:Map2_vis}
\end{figure*}

\begin{figure*}[htb]
    \centering
    \begin{subfigure}{0.2\textwidth}
        \includegraphics[width=\textwidth]{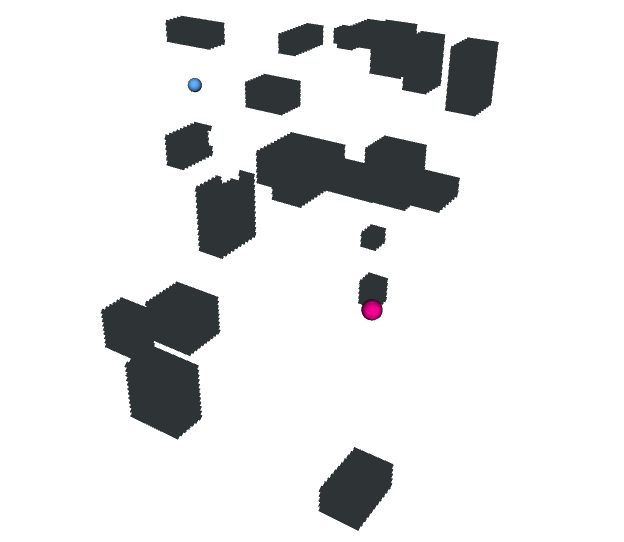}
        \caption{Original map.}
        \label{fig:map3-1}
    \end{subfigure}%
    \begin{subfigure}{0.2\textwidth}
        \includegraphics[width=\textwidth]{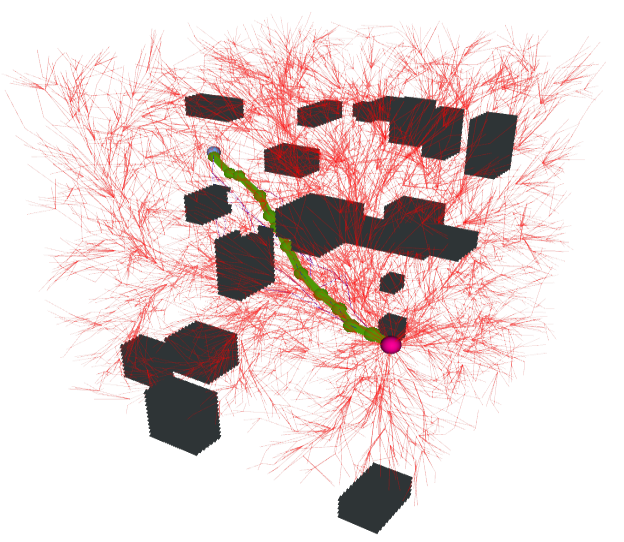}
        \caption{RRT*'s front-end path.}
        \label{fig:map3-2}
    \end{subfigure}%
    \begin{subfigure}{0.2\textwidth}
        \includegraphics[width=\textwidth]{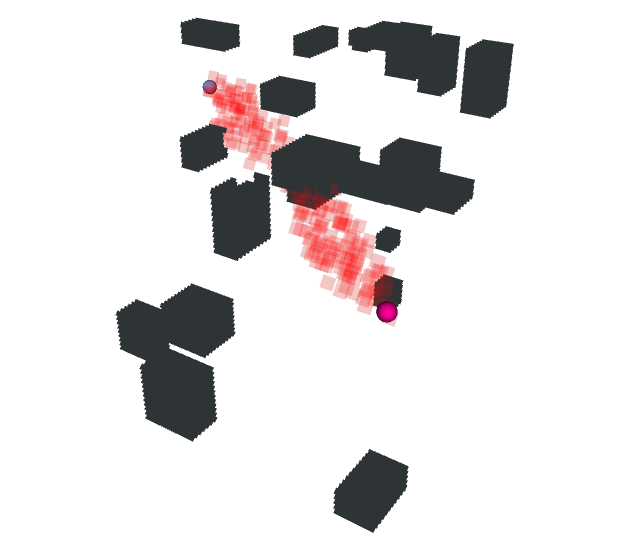}
        \caption{Heuristic region.}
        \label{fig:map3-3}
    \end{subfigure}%
    \begin{subfigure}{0.2\textwidth}
        \includegraphics[width=\textwidth]{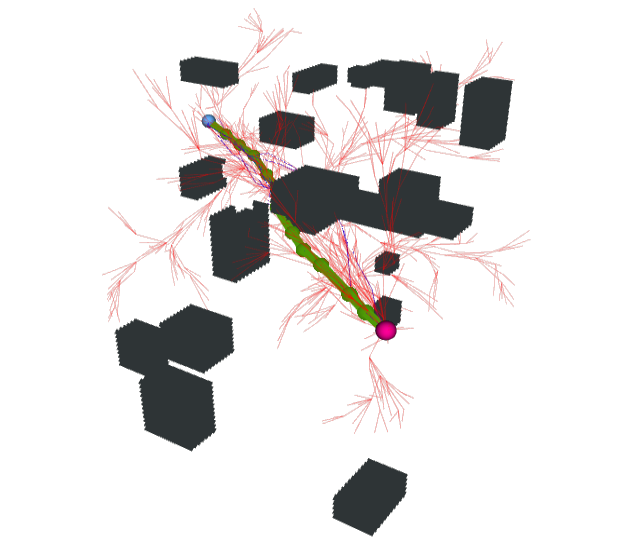}
        \caption{Our front-end path.}
        \label{fig:map3-4}
    \end{subfigure}%
    \begin{subfigure}{0.2\textwidth}
        \includegraphics[width=\textwidth]{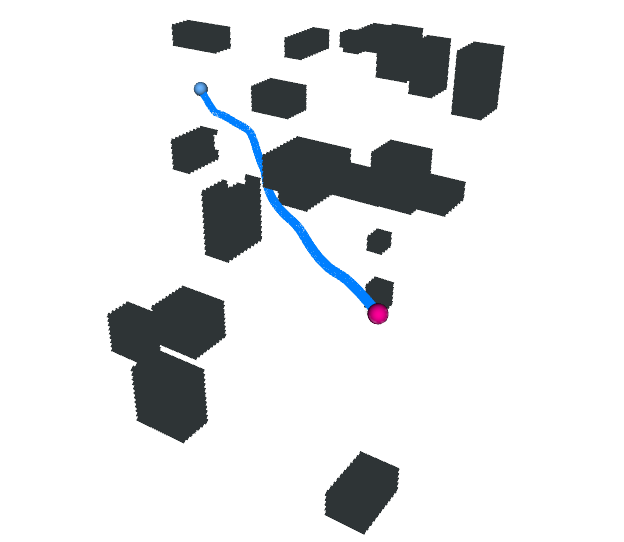}
        \caption{Our final trajectory.}
        \label{fig:map3-5}
    \end{subfigure}
    \caption{The visualization of trajectory planning process: original map, RRT*'s front-end path, heuristic region, our front-end path, our final trajectory (Simulation map3).}
    \label{fig:Map3_vis}
\end{figure*}

The process of the algorithm is illustrated in Alg.~\ref{alg1}. First, the state map $\mathcal{S}$ and 3D environment map $\mathcal{E}$ are shown in Fig.~\ref{fig:map1-1}-\ref{fig:map3-1}. The trained 3D sGAN model will generate a promising region $\mathcal{R}$ as the heuristic [Alg.~\ref{alg1} Lines 1-3] to guide the sampling of RRT*, as shown in Fig.~\ref{fig:map1-3}-\ref{fig:map3-3}. 

\begin{figure}[htb]
    \centering
    \includegraphics[width=0.99\columnwidth]{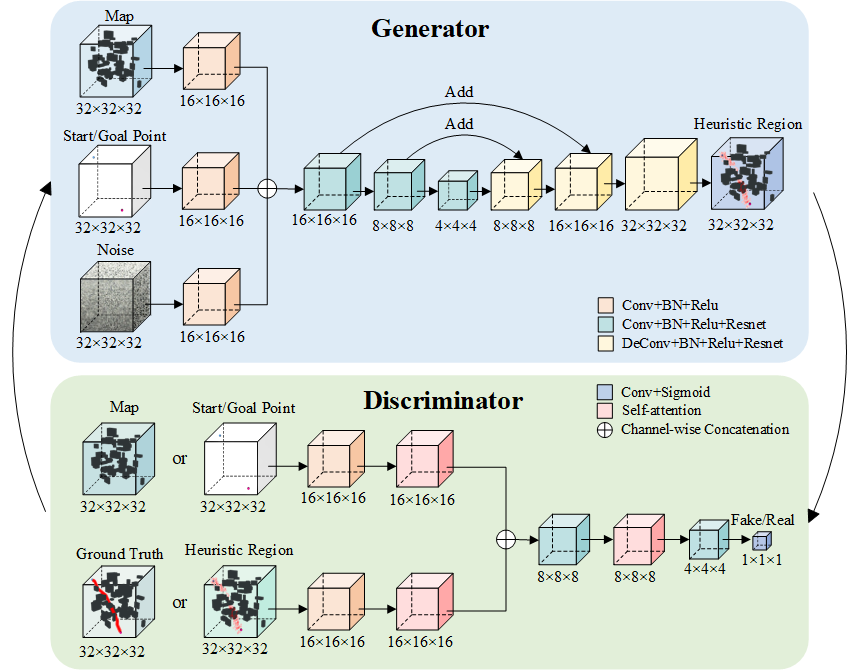}
    \caption{Architecture of the proposed network.}
    \label{fig:sGAN}
\end{figure}

Next, we design a biased sampling strategy [Alg.~\ref{alg1} Lines 5-17]. When $Rand()$ is less than a certain threshold $\mu _{i|i=1,2}$, we will sample in the heuristic promising region, otherwise the whole environment. In the initial stage, we hope to sample as much as possible in the heuristic region to quickly find the initial solution. After finding the initial solution, we need to ensure that the algorithm will not fall into the local optimum and quickly converge to the optimal solution.

Other parts of the algorithm [Alg.~\ref{alg1} Lines 18-26] are similar to the classic RRT* algorithm, and the latest modified rewiring radius method~\cite{solovey2020revisiting} is used at [Alg.~\ref{alg1} Line 21]. The final output will be a tree $G=\left( V_{N},E \right)$, where a set of vertices $V_{N}\subset \mathcal{X} _{\mathrm{free}}$ and edges $E=\left\{ \left( \bar{v}_1,\bar{v}_2 \right) \right\}$ for $\bar{v}_1,\bar{v}_2\in V_{N}$. $G$ records the waypoints and cost of the asymptotically optimal path for subsequent trajectory optimization. The minimum cost obtained by tree query through our algorithm is defined as $C_{\text{N}}$.

\subsubsection{3D Conditional sGAN Model}

Inspired by~\cite{mirza2014conditional}, we use conditional generative adversarial networks to predict heuristic regions. The core idea is to generate new data samples through adversarial training, where the generator $G$ tries to construct a non-linear mapping from the latent space to the output space $l\rightarrow o$ based on certain conditions $h$. sGAN introduces conditional information to provide better control over the generation and improve the generated data's quality. The discriminator $D$ tries to distinguish between real data $gt$ and generated data $o$ in terms of both safety and connectivity. In our setting, the latent variable is random noise satisfying $\mathcal{N} \sim \left( 0,1 \right) $, $o$ is the generator output, and the conditional information $h$ includes 3D environment map $\mathcal{E}$ and state map $\mathcal{S}$.

As shown in Fig.~\ref{fig:sGAN}, the discriminator $D$ takes in either the predicted heuristic map $o$ or the ground truth map $gt$, conditioned on information $\mathcal{E}$ or $\mathcal{S}$. $D$ contains four volumetric convolution layers and connects a sigmoid function at the end to output between $0\left(o\right)$ and $1\left(gt\right)$. To improve the area's connectivity from the start point to the end point while generating a safe heuristic area, we set $\mathcal{L} _{D_{\mathrm{connectivity}}}$ to better locate the state map and $\mathcal{L} _{D_{\mathrm{safety}}}$ to avoid collision between the heuristic region and environment map:
\begin{equation}
\begin{aligned}
    & \mathcal{L} _{D_{\mathrm{connectivity}}} = \mathbb{E} \left[ \log D_{\mathrm{connectivity}}\left( \mathcal{R}|\mathcal{S} \right) \right]\\
    & \quad \quad \qquad \quad + \mathbb{E} \left[ \log \left( 1-D_{\mathrm{connectivity}}\left( G\left( \mathcal{N}|\mathcal{E},\mathcal{S} \right) ,\mathcal{S} \right) \right) \right]. \\
\end{aligned}
\label{eq:connectivity_loss}
\end{equation}
\begin{equation}
\begin{aligned}
    & \mathcal{L} _{D_{\mathrm{safety}}} = \mathbb{E} \left[ \log D_{\mathrm{safety}}\left( \mathcal{R}|\mathcal{E} \right) \right]\\
    & \quad \quad \qquad + \mathbb{E} \left[ \log \left( 1-D_{\mathrm{safety}}\left( G\left( \mathcal{N}|\mathcal{E},\mathcal{S} \right) ,\mathcal{E} \right) \right) \right]. \\
\end{aligned}
\label{eq:safety_loss}
\end{equation}
The generator $G$ extracts noise map $\mathcal{N}$, 3D environment map $\mathcal{E}$, and state map $\mathcal{S}$ at different scales through convolution and then concatenates them together, as shown in Fig.~\ref{fig:sGAN}. Subsequently, we adopt an encoder-decoder structure~\cite{johnson2016perceptual} and add a sigmoid function at the end to generate a promising region heuristic map $\mathcal{R}$. To balance $\mathcal{L} _{D_{\mathrm{safety}}}$ and $\mathcal{L} _{D_{\mathrm{connectivity}}}$ and give similar importance to safety and connectivity, we design the loss function of training $G$ as:
\begin{equation}
\begin{aligned}
    & \mathcal{L} _{G}=\frac{\kappa \mathcal{L} _{D_{\mathrm{safety}}}}{\mathcal{L} _{D_{\mathrm{connectivity}}}+\kappa \mathcal{L} _{D_{\mathrm{safety}}}}\mathbb{E}\left[ \log D_{\mathrm{safety}}\left( G\left( \mathcal{N}|\mathcal{E},\mathcal{S} \right) ,\mathcal{E} \right) \right] \\
    & + \frac{\mathcal{L} _{D_{\mathrm{connectivity}}}}{\mathcal{L} _{D_{\mathrm{connectivity}}}+\kappa \mathcal{L} _{D_{\mathrm{safety}}}}\mathbb{E}\left[ \log D_{\mathrm{connectivity}}\left( G\left( \mathcal{N}|\mathcal{E},\mathcal{S} \right) ,\mathcal{S} \right) \right], \\
\end{aligned}
\label{eq:generator_loss}
\end{equation}
where $\kappa$ is the balance coefficient.

The general objective function of our sGAN model is defined as follows:

\begin{equation}
\begin{aligned}
    & \mathcal{L} _{sGAN}\left( D,G \right) =\mathbb{E}_{o,h}\left[ \log D\left( o|h \right) \right] +\mathbb{E}_{l,h}\left[ \log \left( 1-D\left( G\left( l|h \right) \right) \right) \right]. \\
\end{aligned}
\end{equation}

In Equations~(\ref{eq:connectivity_loss})~(\ref{eq:safety_loss})~(\ref{eq:generator_loss}), we have leveraged the intrinsic properties of conditional GANs with the expectation that the network can learn the characteristics of connectivity and safety during training, but we have not yet introduced any explicit penalties. Therefore, we need to further incorporate connectivity and safety into the final loss function in an explicit and differentiable manner.

Firstly, in order to ensure the continuity of heuristic regions, we introduce a loss term that penalizes non-contiguous regions in the map. This is achieved by calculating the continuity between adjacent points, where a high loss value is incurred if the distance between neighboring points exceeds a certain threshold:
\begin{equation}
    \mathcal{L} _{\text{connectivity}}\left( p \right) =\sum_{\scriptstyle i\in N \atop \scriptstyle j\in \text{Neighbours}(i)}{\phi (|p_i-p_j|-\delta )},
\end{equation}
where $p$ is the point in the heuristic region, $|p_i-p_j|$ denotes the distance between points $i$ and $j$, $N$ is the set of all points within the heuristic region, $\text{Neighbours}(i)$ is the set of neighboring points of point $i$ in the heuristic region, $\phi(x) = \max(0, x)$ is the penalty function, and $\delta$ is a threshold parameter.

Secondly, to reduce the number of obstacle points included in the heuristic region, another loss term is required to penalize these points. We define this loss term as follows:
\begin{equation}
    \mathcal{L} _{\mathrm{safety}}(p,\mathcal{E} )=\underset{i\in N}{\varSigma}\,\,\psi (p_i\cdot e_i),
\end{equation}
where $p_i$ is the probability value of point $i$ in the heuristic region, $e_i$ represents the value of point $i$ in the environmental map $\mathcal{E}$ (1 if $i$ is an obstacle point, otherwise 0), and $\psi (x)=1/\left( 1+e^{-x} \right) $ is the penalty function.

Finally, similar to~\cite{pathak2016context}, we introduce the classic binary cross-entropy (BCE) loss function $\mathcal{L} _{BCE}$ to improve the quality of generation while reducing training difficulty. Therefore, the final objective function is:

\begin{equation}
\begin{aligned}
    \theta ^* &= \alpha _1 \cdot \mathrm{arg}\underset{G}{\min}\underset{D}{\max} \mathcal{L}_{\text{sGAN}}(D,G) + \alpha _2 \cdot \mathcal{L}_{\text{BCE}}(o,gt) \\
    &\qquad + \alpha _3 \cdot \mathcal{L}_{\text{continuity}}(p) + \alpha _4 \cdot \mathcal{L}_{\text{safety}}(p,\mathcal{E}),
\end{aligned}
\end{equation}
where $\alpha _1$, $\alpha _2$, $\alpha _3$ and $\alpha _4$ are weight coefficients.

\begin{algorithm}
	\renewcommand{\algorithmicrequire}{\textbf{Input:}}
	\renewcommand{\algorithmicensure}{\textbf{Output:}}
	\caption{3D-sGAN-RRT*}
	\label{alg1}
	\begin{algorithmic}[1]
        \REQUIRE $x_{\mathrm{init}}, x_{\mathrm{goal}}, Map$
 	\ENSURE $G$
		\STATE $V_{N}\gets \left\{ x_{\mathrm{init}} \right\} , E\gets \emptyset , G=\left( V_{N},E \right)$
		\STATE $H\gets 3D sGAN\left( x_{\mathrm{init}}, x_{\mathrm{goal}}, \mathcal{E} \right)$
        \STATE $H\gets Filter\left( H \right)$

        \FOR {$i=1\cdots N$}
            \IF{\textbf{FirstGoalFound}()}
                \IF {$Rand\left(\right) <\mu_1$}
                    \STATE $x_{\mathrm{rand}}\gets Nonuniform\left( H \right)$
                \ELSE
                    \STATE $x_{\mathrm{rand}}\gets Uniform\left( Map \right) $
                \ENDIF
            \ELSE
                \IF {$Rand\left( \right) <\mu_2$}
                    \STATE $x_{\mathrm{rand}}\gets Nonuniform\left( H \right)$
                \ELSE
                    \STATE $x_{\mathrm{rand}}\gets Uniform\left( Map \right) $
                \ENDIF
            \ENDIF
            
            \STATE $x_{\mathrm{nearest}}\gets Nearest\left( G=\left( V_{N},E \right) , x_{\mathrm{rand}} \right)$
            \STATE $x_{\mathrm{new}}\gets Steer\left( x_{\mathrm{nearest}}, x_{\mathrm{rand}} \right)$
            \IF {$CollisionFree\left( x_{\mathrm{nearest}}, x_{\mathrm{new}} \right)$}
                \STATE $G\longleftarrow Extend\left( G, x_{\mathrm{new}} \right)$
                \STATE $Rewire\left(  \right)$
                \IF {$x_{\mathrm{new}}\in \mathcal{X}_{\mathrm{goal}}$}
                    \STATE Return $G$
                \ENDIF
            \ENDIF
        \ENDFOR
        \STATE Return $failure$
        
	\end{algorithmic}  
\end{algorithm}

\subsection{Trajectory Planning Back-end: Optimal Trajectory Generation}
\label{sec:Back-end}

\subsubsection{Minimum Control Effort Trajectory Generation}

In the previous section, we obtain high-quality initial waypoints and paths. In this part, we optimize the initial path and generate final trajectories that satisfy the corresponding constraints. (\ref{eqn:minco1}) is the general form of multi-stage unconstrained minimization control effort trajectory optimization problem. In our UAV trajectory generation tasks, the focus is on minimizing the first derivative (jerk) and the second derivative (snap) of the acceleration, which are related to angular velocity (e.g., good for visual tracking) and differential thrust (e.g., saves energy), respectively.

To solve this problem faster, we adopt the necessary and sufficient conditions of optimality proposed in~\cite{wang2022geometrically}. This condition tells us how to directly construct the unique and optimal trajectory under all possible settings of $d_i, \overline{\gamma}_i, t_i$. 

The optimal $\gamma^*\left( t \right) :\left[ t_{i-1},t_i \right] \mapsto \mathbb{R}^m$ is a polynomial of degree $2s - 1$ for every $1\leqslant i\leqslant M$. For an $m$-dimensional ($m=3$ in our 3D case) trajectory, its $i$-th segment can be represented by a $2s-1$ ($s=3$ for minimum jerk and $s=4$ for minimum snap) order polynomial:

\begin{equation}
    p_i\left( t \right) =c_{i}^{T}\rho \left( t-t_{i-1} \right) , t\in \left[ t_{i-1}, t_i \right], 
\end{equation}
where $\rho \left( x \right) =\left( 1,x,...,x^{2s-1} \right) ^T$ is the polynomial basis and $c_i\in \mathbb{R}^{2s\times m}$ are the coefficients:
\begin{equation}
    c=\left( c_{1}^{T},...,c_{M}^{T} \right) ^T, T=\left( T_1,...,T_M \right) ^T.
\end{equation}

For the trajectory generation problem characterized by polynomial splines, the linear constraints can be divided into two categories. The derivative constraint (\ref{eqn:minco2}) specifies the derivative values of each order at the start and end points of each trajectory. The continuity constraint (\ref{eqn:minco3}) specifies that the values of derivatives of each order at the junction of the two trajectories before and after are the same. When we apply the optimality condition, we can obtain the following linear equation about the coefficient matrix:
\begin{equation}
    Ac=b,
\label{coeff}
\end{equation}
where $A\in \mathbb{R}^{2Ms\times 2Ms}$ is a nonsingular banded sparse matrix for any positive time allocation and $b\in \mathbb{R}^{2Ms\times m}$. The construction of matrix $A$ and $b$ is demonstrated in [Alg.~\ref{alg2} Lines 3-7]: First, $F_0\in \mathbb{R}^{s\times 2s}$ and $C_0\in \mathbb{R}^{s\times m}$ are constructed based on the start-point conditions; then, $F_i\in \mathbb{R}^{2s\times 2s}$, $C_i\in \mathbb{R}^{d_i\times m}$, and $E_i\in \mathbb{R}^{2s\times 2s}$ are formed following the intermediate-points conditions; finally, $E_M\in \mathbb{R}^{s\times 2s}$ and $C_M\in \mathbb{R}^{s\times m}$ are built according to the end-point conditions. (\ref{coeff}) can be solved with $O\left( M \right)$ time and space complexity to get $c$ using banded PLU factorization~\cite{horn2012matrix}.

\begin{algorithm}
	\renewcommand{\algorithmicrequire}{\textbf{Input:}}
	\renewcommand{\algorithmicensure}{\textbf{Output:}}
	\caption{Trajectory Generation (Rest-to-Rest)}
	\label{alg2}
	\begin{algorithmic}[1]
        \REQUIRE $\mathcal{W} \gets G.$\textbf{GetWaypoints}(), Derivative order $s$
 	\ENSURE Trajectory $\mathcal{P}=p_i\left( t \right) _{i=1}^{M}, \forall t\in \left[ t_{i-1},t_i \right)$
		\STATE $M$, $b$, Time vector $\mathcal{T}\gets$\textbf{Initialization}($\mathcal{W}, s$)
        \STATE $\mathcal{T} \gets$\textbf{TrapezoidalTimeAllocation}($\mathcal{W}$)
        \STATE $M.$\textbf{AddValue}($F_0$), $b.$\textbf{AddValue}($C_0$)
        \FOR {$i=0,...,\left( \mathcal{T}.\textbf{GetSize}() -1 \right) *2s$}
            \STATE $M.$\textbf{AddValue}($F_i, E_i$), $b.$\textbf{AddValue}($C_i$)
        \ENDFOR        
        \STATE $M.$\textbf{AddValue}($C_M$), $b.$\textbf{AddValue}($C_M$)  

        \STATE Flag $= 0$
        \FOR {$i=1,...,M, p_i\left( t \right)$}
            \IF {\textbf{CheckCollisionFree}$\left( p_i\left( t \right) \right)$}
                \STATE continue
            \ELSE
                \STATE $\mathcal{W}$.\textbf{AddHalfwayPoint}($p_i\left( t \right)$)
                \STATE Flag $= 1$
            \ENDIF   
        \ENDFOR
        \IF{! Flag}    
            \STATE Return $\mathcal{P}$
        \ELSE
            \STATE Back to line 1
        \ENDIF
	\end{algorithmic}  
\end{algorithm}

\begin{figure*}[htb]
    \centering
    \begin{subfigure}{0.24\textwidth}
        \includegraphics[width=\textwidth]{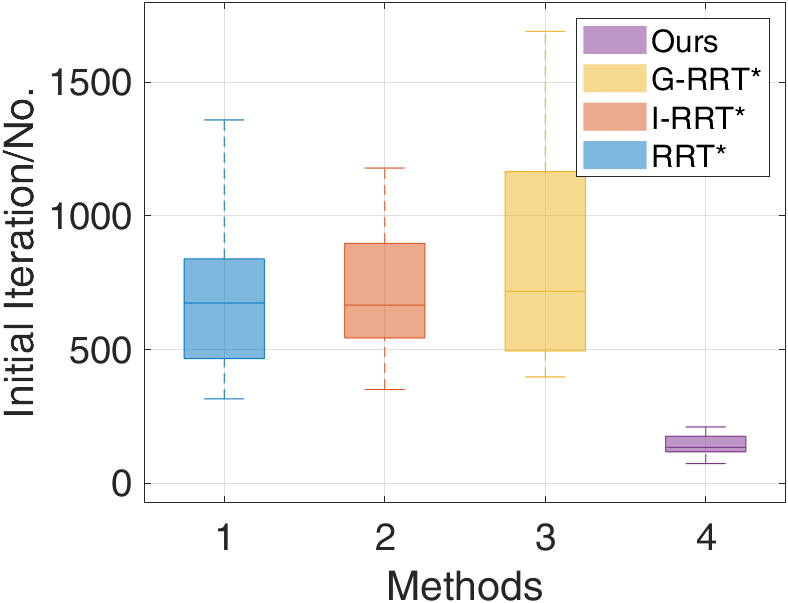}
        \caption{}
        \label{fig:BigMap1_a}
    \end{subfigure}
    \begin{subfigure}{0.24\textwidth}
        \includegraphics[width=\textwidth]{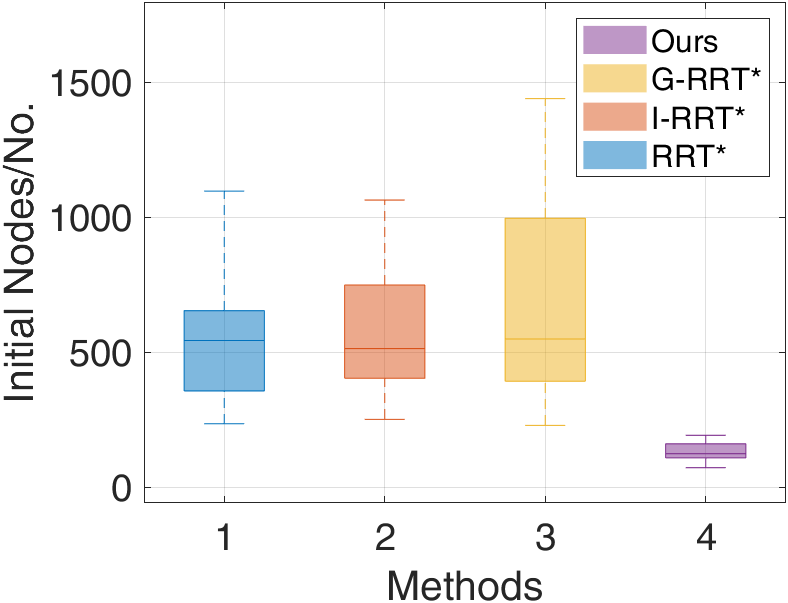}
        \caption{}
        \label{fig:BigMap1_b}
    \end{subfigure}
    \begin{subfigure}{0.235\textwidth}
        \includegraphics[width=\textwidth]{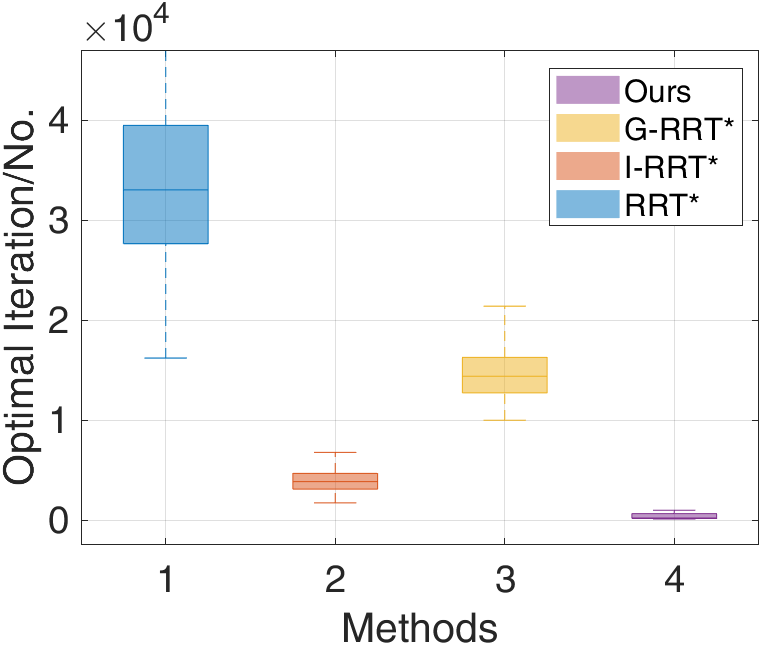}
        \caption{}
        \label{fig:BigMap1_c}
    \end{subfigure}
    \begin{subfigure}{0.235\textwidth}
        \includegraphics[width=\textwidth]{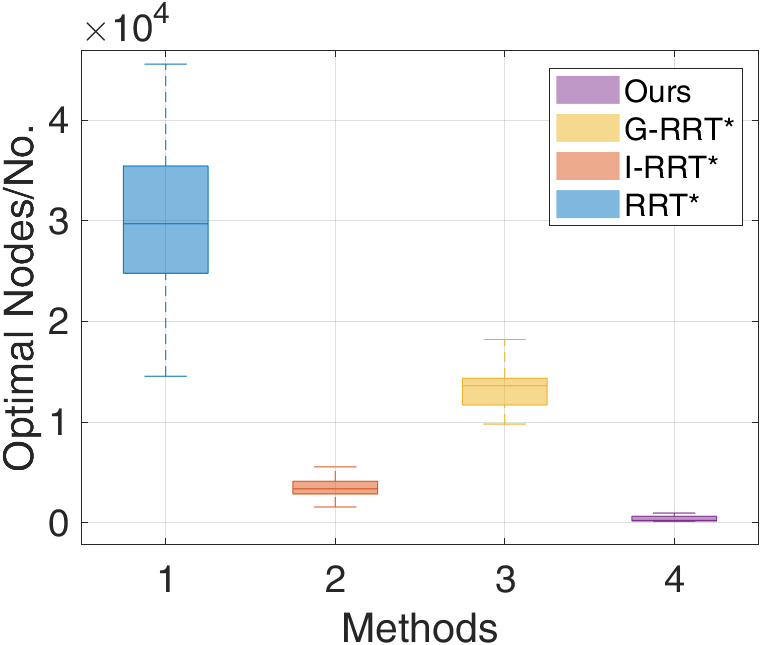}
        \caption{}
        \label{fig:BigMap1_d}
    \end{subfigure}
    \caption{Comparison between our method and three other SOTA methods (RRT*, I-RRT*, G-RRT*) in terms of iteration and nodes in the initial phase and asymptotically optimal phase (Simulation map1).}
    \label{fig:Map1}
\end{figure*}

\begin{figure*}[htb]
    \centering
    \begin{subfigure}{0.24\textwidth}
        \includegraphics[width=\textwidth]{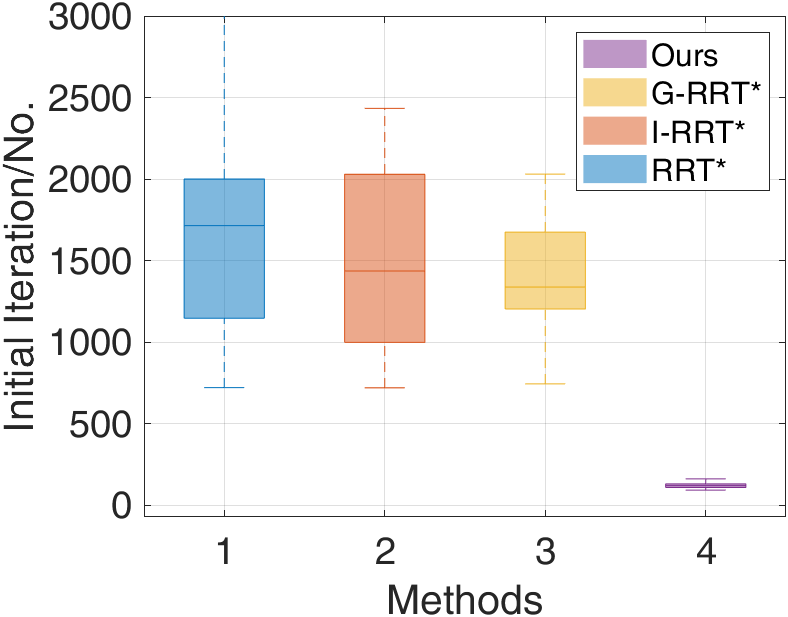}
        \caption{}
        \label{fig:BigMap2_a}
    \end{subfigure}
    \begin{subfigure}{0.24\textwidth}
        \includegraphics[width=\textwidth]{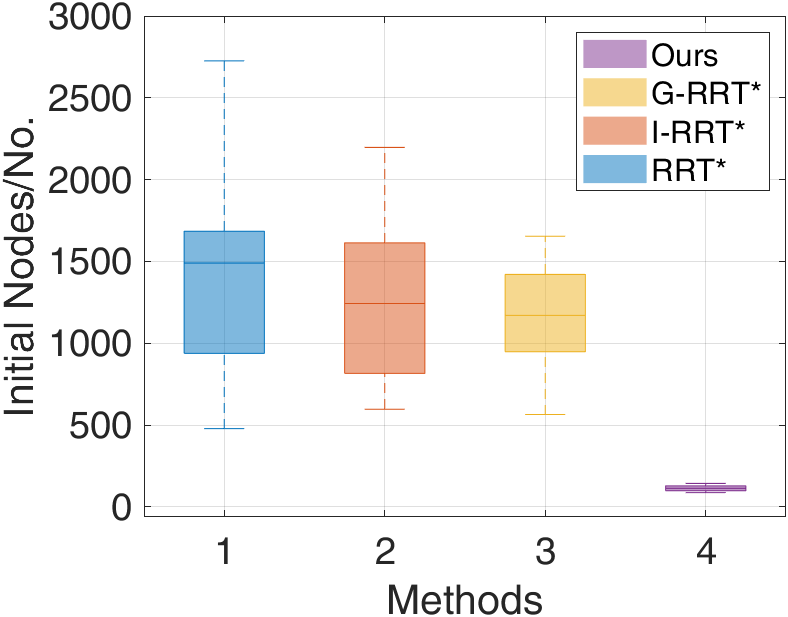}
        \caption{}
        \label{fig:BigMap2_b}
    \end{subfigure}
    \begin{subfigure}{0.24\textwidth}
        \includegraphics[width=\textwidth]{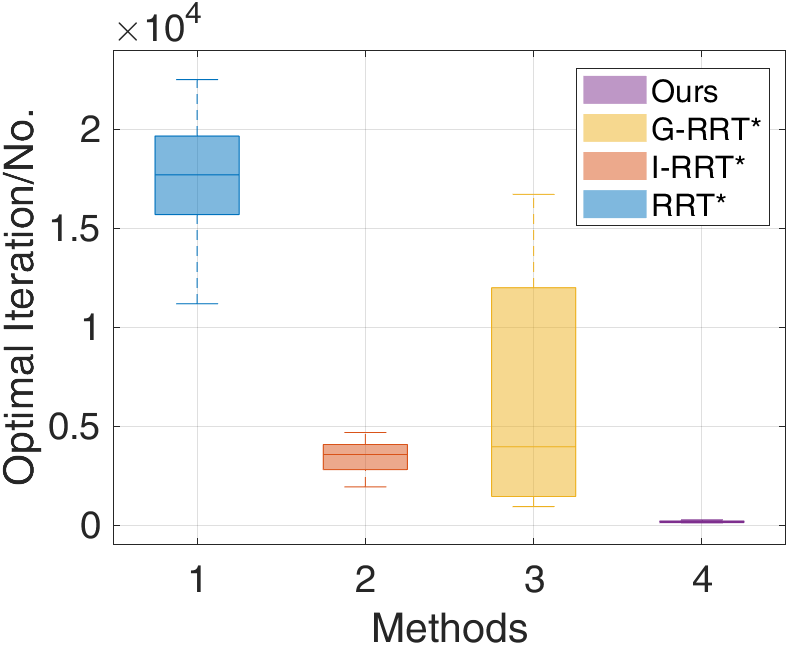}
        \caption{}
        \label{fig:BigMap2_c}
    \end{subfigure}
    \begin{subfigure}{0.24\textwidth}
        \includegraphics[width=\textwidth]{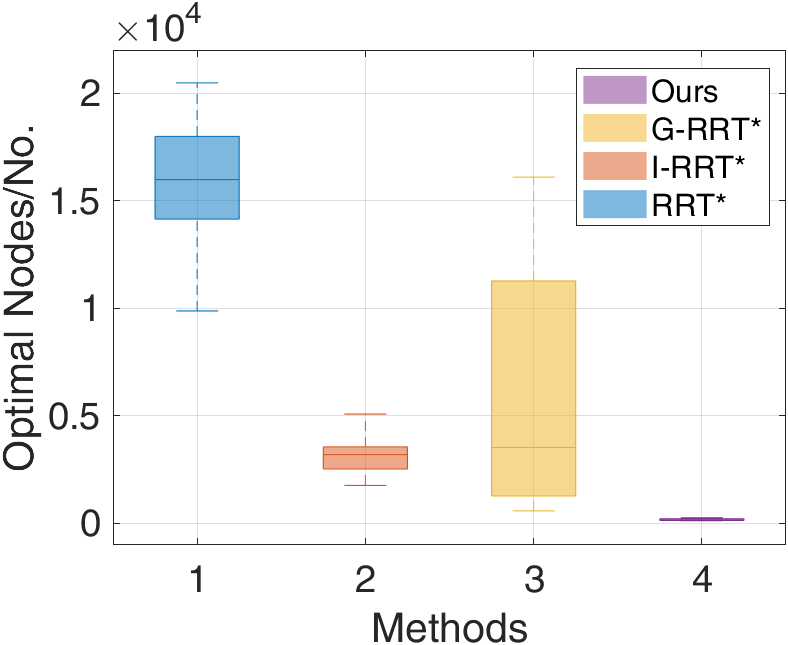}
        \caption{}
        \label{fig:BigMap2_d}
    \end{subfigure}
    \caption{Comparison between our method and three other SOTA methods (RRT*, I-RRT*, G-RRT*) in terms of iteration and nodes in the initial phase and asymptotically optimal phase (Simulation map2).}
    \label{fig:Map2}
\end{figure*}

\begin{figure*}[htb]
    \centering
    \begin{subfigure}{0.24\textwidth}
        \includegraphics[width=\textwidth]{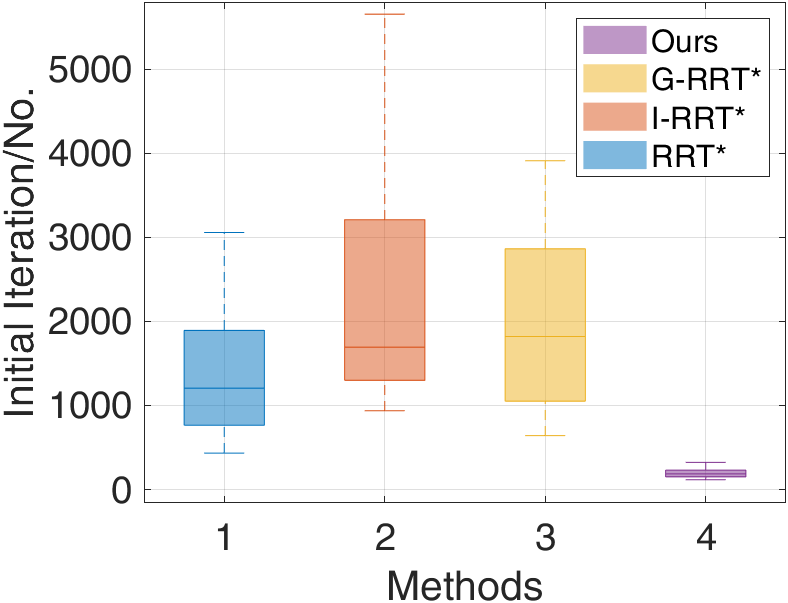}
        \caption{}
        \label{fig:BigMap3_a}
    \end{subfigure}
    \begin{subfigure}{0.24\textwidth}
        \includegraphics[width=\textwidth]{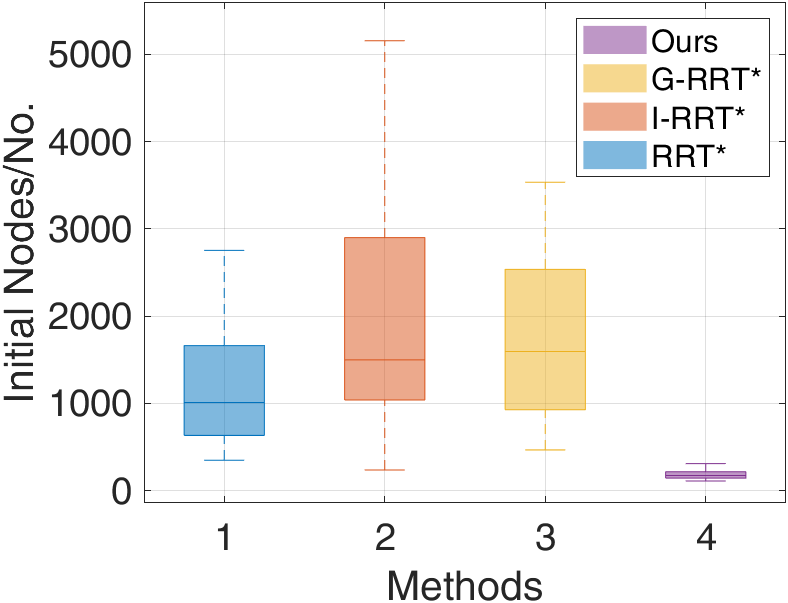}
        \caption{}
        \label{fig:BigMap3_b}
    \end{subfigure}
    \begin{subfigure}{0.24\textwidth}
        \includegraphics[width=\textwidth]{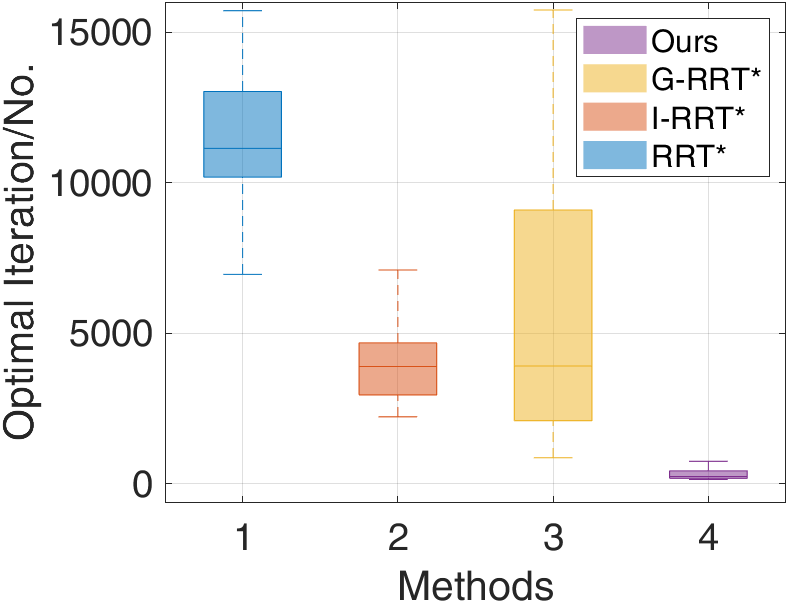}
        \caption{}
        \label{fig:BigMap3_c}
    \end{subfigure}
    \begin{subfigure}{0.24\textwidth}
        \includegraphics[width=\textwidth]{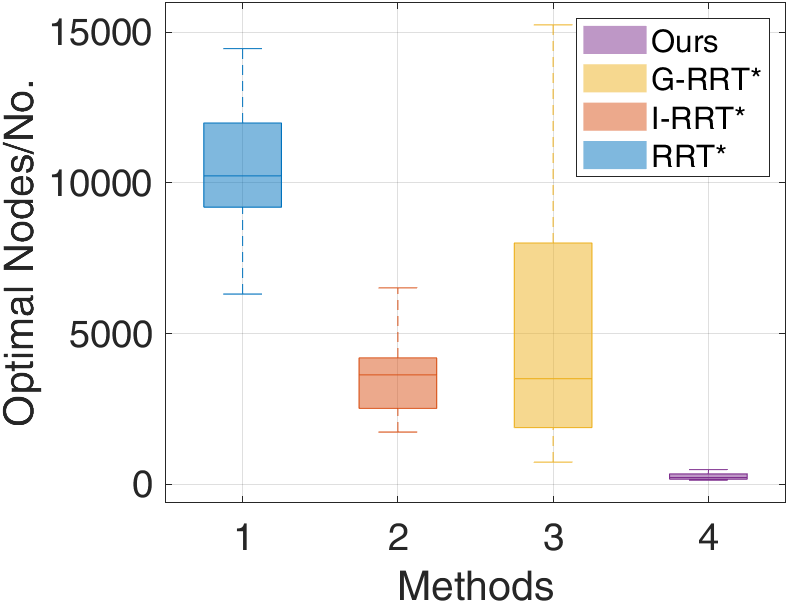}
        \caption{}
        \label{fig:BigMap3_d}
    \end{subfigure}
    \caption{Comparison between our method and three other SOTA methods (RRT*, I-RRT*, G-RRT*) in terms of iteration and nodes in the initial phase and asymptotically optimal phase (Simulation map3).}
    \label{fig:Map3}
\end{figure*}

\subsubsection{Collision-free Guarantee and Time Allocation}

Although obstacle inflation can be added in the front-end path search, the smooth trajectory obtained after the back-end trajectory optimization may still intersect with obstacles in a particular section. Therefore, similar to~\cite{bry2015aggressive}, a simple but effective solution is to insert a new waypoint in the middle of the trajectory segment with collision and then re-do the time allocation and trajectory optimization to obtain the final collision-free trajectory iteratively, which is shown in [Alg.~\ref{alg2} Lines 8-20].

In (\ref{eqn:minco1}), it is necessary to allocate a reasonable time period to each trajectory in advance. Based on~\cite{shomin2014fast}, this paper adopts a trapezoidal time profile to produce the time duration. Specifically, the UAV will first increase to the maximum speed with the maximum acceleration, cruise at the maximum speed, and then decrease to $0$ with the negative value of the maximum acceleration. Thus the time profile $\mathcal{T}$ can be calculated as follows:

\begin{equation}
\begin{gathered}
    \mathcal{T} =\left\{ t_i=\begin{cases}
    	2\sqrt{d_i/a_{\max}}\,\,\left( if\,\,d_i<2d_{\max} \right)\\
    	2t+\left( d_i-2d_{\max} \right) /v_{\max}\,\,\left( else \right)\\
    \end{cases}|i=1,...,M \right\} \\
    t=v_{\max}/a_{\max}, \, d={a_{\max}}^2/2,
\end{gathered}
\end{equation}
where $v_{\max}$ is the maximum velocity, $a_{\max}$ is the maximum acceleration, $t_i$ and $d_i$ are the time and distance of the $i$-th path segment, $t_m$ and $d_m$ are the possible time and distance during acceleration (deceleration).

\subsection{Probabilistic Completeness and Optimality Analysis}
\label{sec:analysis}

Combining the previous content, the entire algorithm flow is shown in Alg.~\ref{alg3}. In [Lines 5-17] of Alg.~\ref{alg1}, regardless of whether the initial solution is found, 3D-sGAN-RRT* will sample the global map with probability ($1-\mu _{i|i=1,2}$). This shows that when the area of sampling points goes to infinity, the position at (around) the target point will always be sampled.\\

\textit{Lemma 1 (Probabilistic Completeness of MINER-RRT*):} The MINER-RRT* algorithm is probabilistically complete, i.e., for any robustly feasible~\cite{karaman2011sampling} planning problem, the following holds:
\begin{equation}
    \lim_{N\rightarrow \infty} \mathbb{P}\left( V_{N}\cap \mathcal{X} _{\mathrm{goal}}\ne \emptyset \right) =1.
\end{equation}
where $V_{N}$ is the set of vertices after $N$ iterations and $\mathcal{X} _{\mathrm{goal}}$ is the goal region.

\begin{proof}
We first define $V_{\text{N}}^{\text{ALG}}$ as the final set of vertices for another certain algorithm (ALG) when the number of sampling nodes is $N$. According to~\cite{lavalle1998rapidly}~\cite{karaman2011sampling}, both RRT and RRT* enjoy probabilistic completeness. When $N\rightarrow \infty$, we can get $V_{\text{N}}=V_{\text{N}}^{\text{RRT}^*}=V_{\text{N}}^{\text{RRT}}$. Since MINER-RRT* can also obtain a connected graph, it possesses the same probabilistic completeness property as RRT and RRT*.
\end{proof}

\textit{Lemma 2 (Asymptotic Optimality of MINER-RRT*):} The MINER-RRT* algorithm is asymptotically optimal, i.e., for any path planning triplet $\left( \mathcal{X} _{\mathrm{free}}, x_{\mathrm{init}}, \mathcal{X} _{\mathrm{goal}} \right)$, optimal cost $c\left( \zeta ^* \right)$ and minimum cost $C_{\text{N}}$ obtained by tree query through our algorithm, the following holds:
\begin{equation}
    P\left( \left\{ \lim_{N\rightarrow \infty} \sup C_{\text{N}} = c\left( \zeta^* \right) \right\} \right) = 1,
\end{equation}
if we let the searching radius $\eta$ in $Extend\left( G, x_{\mathrm{new}} \right)$ in Alg.~\ref{alg1} [Line 21] satisfy the following condition~\cite{karaman2011sampling}:
\begin{equation}
    \eta >\left( 2\left( 1+\frac{1}{m} \right) \right) ^{\frac{1}{m}}\left( \frac{\mathcal{M} \left( \mathcal{X} _{\mathrm{free}} \right)}{\xi _d} \right) ^{\frac{1}{m}},
\end{equation}
where $\mathcal{M} \left( \mathcal{X} _{\mathrm{free}} \right)$ is the Lebesgue measure of $\mathcal{X} _{\mathrm{free}}$, $\xi _d$ is the volume of the unit ball.

\begin{proof}
Since MINER-RRT* is an improvement to the sampling strategy and will not change the $Extend\left( G, x_{\mathrm{new}} \right)$ and $Rewire()$ processes of RRT* itself, our method will inherit the characteristics of RRT* and possesses asymptotic optimality according to the theorem 38 in~\cite{karaman2011sampling}, provided that the searching radius satisfies the above conditions.
\end{proof}

According to~\cite{wang2022geometrically}, the unique optimal trajectory for the linear quadratic problem of minimizing unconstrained control effort trajectories with multiple stages is in the form of (\ref{eqn:minco1})-(\ref{eqn:minco5}). The optimal solution, denoted as $\gamma^*\left( t \right)$, is a $2s-1$ degree polynomial and is continuous and differentiable up to the $\overline{d_i}-1$-th order at each $t_i$ ($\overline{d_i}=2s-d_i$). Additionally, the trajectory satisfies both intermediate and boundary conditions. Based on the description provided, our hierarchical trajectory planning framework is guaranteed to find the desired polynomial trajectory under the above-given settings.

\begin{algorithm}
	\renewcommand{\algorithmicrequire}{\textbf{Input:}}
	\renewcommand{\algorithmicensure}{\textbf{Output:}}
	\caption{MINER-RRT*}
	\label{alg3}
	\begin{algorithmic}[1]
        \REQUIRE $x_{\mathrm{init}}, x_{\mathrm{goal}}, Map$, Derivative order $s$
 	\ENSURE Optimal trajectory $\mathcal{P} =p_i\left( t \right) _{i=1}^{M}, \forall t\in \left[ t_{i-1},t_i \right)$
		\STATE $G\gets \mathbf{3D-sGAN-RRT^*} \left( x_{\mathrm{init}}, x_{\mathrm{goal}}, Map \right)$
        \STATE $\mathcal{W} \gets G.$\textbf{GetWaypoints}()
        \STATE $\mathcal{P} \gets $ \textbf{Trajectory Generation}($\mathcal{W}$, $s$)
        \STATE Return $\mathcal{P}$
	\end{algorithmic}  
\end{algorithm}




\section{EXPERIMENTAL RESULTS AND ANALYSIS}

\subsection{Training Details}

\subsubsection{Data Generation}

We create a dataset comprising 19,493 samples from successful path-finding scenarios in 3D cluttered environments for training and validation. Concurrently, we also randomly generate maps that the model has not encountered during the training phase to serve as a test set, comprising 605 samples. Some unseen test samples are shown in Fig.~\ref{fig:Map1_vis}-\ref{fig:Map3_vis} (a), where a grid map represents the 3D environment $\mathcal{E}$, with obstacle blocks and start/goal point in it. 

We apply the A* algorithm to obtain the ground truth path for each environment and then expand the path using dilation for better neural network recognition. Specifically, to reinforce key features and enhance robustness during convolution and pooling operations, certain points within the 3D images, such as the start point, end point, and each point along the A* trajectory path, are dilated into $3\times3\times3$ cubes centered on those points.

\subsubsection{Training Process}

We train the proposed sGAN model on 19493 sets of 3D images using Python 3.7.16 with Pytorch 1.12.1 on NVIDIA A40-12Q. The model is trained for 50 epochs and the training time is 2.3 hours. We employ the Adam optimizer, setting parameters $\beta _1$ and $\beta _2$ to 0.5 and 0.99, respectively. The batch size is 32, and the learning steps of $D$ and $G$ are set to 1e-6 and 1e-5.

\subsection{Simulation Experiment}

\subsubsection{Environment Set-up}

We build a Robot Operating System (ROS)-based simulation platform on the Ubuntu 18.04 platform. Maps are built in advance and stored as occupancy grids for obstacle collision detection using Point Cloud Library~\cite{rusu20113d}. All methods are implemented in C++14 using the Eigen package and do not rely on any specific acceleration.

\subsubsection{Evaluation Index}

We select indicators of different dimensions to verify the effectiveness and efficiency of our method. Firstly, the front-end path-finding part is divided into the initial stage and the optimal stage. The initial stage ranges from the beginning of the planner execution to finding the initial path solution, while the optimal stage ranges from finding the initial path solution to finding the asymptotically optimal path solution. `Iteration' is the number of loops executed by the algorithm, `Nodes' is the number of sampling points added to the tree that are valid for all loops, `Cost' is expressed in Euclidean distance, and `Time' is measured in milliseconds. Secondly, in the back-end trajectory optimization and overall trajectory planning, time measured in milliseconds is used as the evaluation index.

\subsubsection{Front-end Comparison}

We first conduct Monte Carlo experiments in different environments and then compare the proposed method with the baseline method (RRT*) and two SOTA heuristic-based RRT* methods (I-RRT* and G-RRT*). Each method is executed 50 times to evaluate its performance. We divide the comparison results into an initial stage and an optimal stage. The `Iteration' and `Time' metrics reflect the time cost of each method, `Nodes' indicates the memory usage, and `Cost' represents the length of the path found by each method.

Fig.~\ref{fig:map1-3}-\ref{fig:map3-3} demonstrate the visualization of heuristic promising regions predicted by the neural network under various random start/end points and unseen environments qualitatively. Tab.~\ref{tab:rate} quantitatively demonstrates the superiority of our method and the existing learning-based heuristic generation method in connectivity rate and safety rate. A connected heuristic region means that a path connecting the starting point to the end point can be directly sampled from the heuristic region only, and a safe heuristic region means that the heuristic region does not contain obstacles that exceed the size of one grid. These above results clearly indicate that our proposed approach can predict high-quality heuristics with satisfying safety and connectivity, which can be efficiently used for sampling in RRT* algorithms.

Fig.~\ref{fig:Map1}-\ref{fig:Map3} and Tab.~\ref{tab:table1} provide a visual representation and comparative statistical analysis of the different methods, respectively. In the initial stage, our method efficiently samples within the heuristic region, thereby quickly identifying a high-quality initial solution. In contrast, the other three methods generate a considerable amount of ineffective samples due to their approach of relatively uniform sampling across the global space. Consequently, our method significantly reduces both the time and memory costs during the initial stage compared to the other methods.

Although RRT*-based algorithms have proven asymptotic optimality, it is often the case in practical robotic tasks that a high-quality solution, close to the optimal one, can be used without excessive iterations. Therefore, we define an optimal solution as one where the algorithm finds a path that meets a pre-set cost value. During the path refining stage, as illustrated in Fig.~\ref{fig:Map1}-\ref{fig:Map3}, our method requires significantly less time and memory to find the asymptotically optimal path compared to other methods. This is due to the generation of high-quality initial solutions by our method and the guidance provided by the heuristic promising region. Furthermore, while traditional RRT* continuously samples uniformly on a global scale, I-RRT* and G-RRT* methods guide the sampling process by respectively constructing $L^2$-informed sets and local subsets. However, if the initial solution is of low quality, the ellipsoid generated will contain a lot of useless space. This can consequently lead to a slower convergence rate.

In summary, as can be observed from Fig.~\ref{fig:Map1}-\ref{fig:Map3} and Tab.~\ref{tab:table1}, our method is superior in terms of time and memory costs. In addition, our method has a high degree of robustness, which ensures the success of the robot's task execution in a single run.

\begin{table}[h]
    \caption{Comparison Results of Heuristic Quality}
    \begin{center}
    \scriptsize  
    \resizebox{.5\textwidth}{!}{
        \renewcommand\arraystretch{1.2}
        \begin{tabular}{c|cc}
        \hline
            Methods & Connectivity Rate & Safety Rate \\ 
        \hline
            Li \textit{et al.}~\cite{li2021efficient} & 91.53\% & 91.90\% \\
            Ma \textit{et al.}~\cite{ma2021conditional} & 91.80\% & 92.40\% \\
            \textbf{Ours} & \textbf{96.30\%} & \textbf{97.10\%} \\ 
        \hline
        \end{tabular}
    }
    \end{center}
    \label{tab:rate} 
\end{table}

\begin{table}[h]
\caption{Front-end Comparison: Different Heuristic-based RRT* Algorithms\label{tab:table1}}
\centering
\renewcommand{\arraystretch}{1.3}
\begin{threeparttable} 
\resizebox{\linewidth}{!}{
\begin{tabular}{p{0.05cm} c | c c c c c c c}
\hline
\multirow{2}{*}{No.} & \multirow{2}{*}{Method} & \multirowcell{2}{Initial\\Iteration} & \multirowcell{2}{Initial\\Nodes} & \multirowcell{2}{Initial\\Cost} & \multirowcell{2}{Initial\\Time} & \multirowcell{2}{Optimal\\Iteration} & \multirowcell{2}{Optimal\\Nodes} & \multirowcell{2}{Optimal\\Time}\\
& & & & & & & \\
\hline
\multirow{4}{*}{1} & RRT* & 737.0 & 590.4 & 99.4 & 2.1 & 34007.9 & 30579.4 & 491.1 \\
                   & I-RRT* & 899.1 & 746.2 & 100.5 & 2.5 & 4118.1 & 3592.4 & 31.9 \\
                   & G-RRT* & 842.2 & 685.9 & 102.9 & 2.8 & 14276.4 & 12995.9 & 102.4 \\
                   & \textbf{Ours} & \textbf{156.9} & \textbf{146.9} & \textbf{82.4} & \textbf{1.2} & \textbf{437.9} & \textbf{415.4} & \textbf{5.3}\\
                   
\hline
\multirow{4}{*}{2} & RRT* & 1729.1 & 1438.1 & 101.4 & 7.3 & 17095.8 & 15549.6 & 135.0\\
                   & I-RRT* & 1536.8 & 1286.1 & 102.6 & 6.2 & 3816.5 & 3274.0 & 27.1\\
                   & G-RRT* & 1503.5 & 1251.5 & 102.0 & 5.0 & 6160.9 & 5673.8 & 45.4\\
                   & \textbf{Ours} & \textbf{122.4} & \textbf{113.5} & \textbf{84.7} & \textbf{1.2} & \textbf{221.7} & \textbf{210.4} & \textbf{3.1}\\

\hline
\multirow{4}{*}{3} & RRT* & 1379.0 & 1174.0 & 99.6 & 7.0 & 11480.1 & 10499.2 & 88.3\\
                   & I-RRT* & 2192.7 & 1882.3 & 99.3 & 10.2 & 3930.5 & 3519.0 & 26.5\\
                   & G-RRT* & 1937.9 & 1706.8 & 98 & 9.9 & 5680.8 & 5294.3 & 44.3\\
                   & \textbf{Ours} & \textbf{209.7} & \textbf{198.5} & \textbf{81.9} & \textbf{2.0} & \textbf{380.7} & \textbf{314.2} & \textbf{4.6}\\
\hline
\end{tabular}
}
\begin{tablenotes} 
\item[No.: Three different map environments in simulation.] 
\item[Unit Description: Iteration \& Nodes/No., Cost/m, Time/ms.] 
\item[RRT*: Benchmark RRT* algorithm~\cite{karaman2011sampling}.]
\item[I-RRT*: Informed RRT* algorithm~\cite{gammell2018informed}.]
\item[G-RRT*: Guided RRT* algorithm~\cite{scalise2023guild}.]
\item[The asymptotic optimal cost of all methods is the same, so it is no longer listed in the table.]
\end{tablenotes} 
\end{threeparttable} 
\end{table}

\begin{table*}[!t]
\caption{Trajectory Planning Framework Runtime Comparison\label{tab:table2}}
\centering
\renewcommand{\arraystretch}{1.3} 
\begin{tabular}{ c | c c c }
\cline{1-4}
  Method (Initial Time (ms) + Min. Jerk/Snap Time (ms)) & Map 1 (Sim.)  & Map 2 (Sim.)& Map 3 (Sim.)\\
\cline{1-4}
  RRT* + Quadratic Programming~\cite{mellinger2011minimum} & 2.10 + 0.550/1.445 & 7.31 + 0.375/0.597 & 7.01 + 0.464/1.023 \\
  RRT* + Closed-form Solution~\cite{bry2015aggressive}\cite{oleynikova2020open} & 2.10 + 2.618/4.390 & 7.31 + 1.809/2.821 & 7.01 + 2.575/4.839 \\
  RRT* + Optimal Condition~\cite{wang2022geometrically} & 2.10 + 0.0023/0.0043 & 7.31 + 0.0022/0.0040 & 7.01 + 0.0024/0.0045 \\
  \textbf{MINER-RRT* (Ours)} & \textbf{1.29 + 0.0023/0.0044} & \textbf{1.20 + 0.0021/0.0039} & \textbf{2.08 + 0.0023/0.0043} \\
\cline{1-4}
\end{tabular}
\end{table*}

\begin{figure}[htb]
    \centering
    \includegraphics[width=0.99\columnwidth]{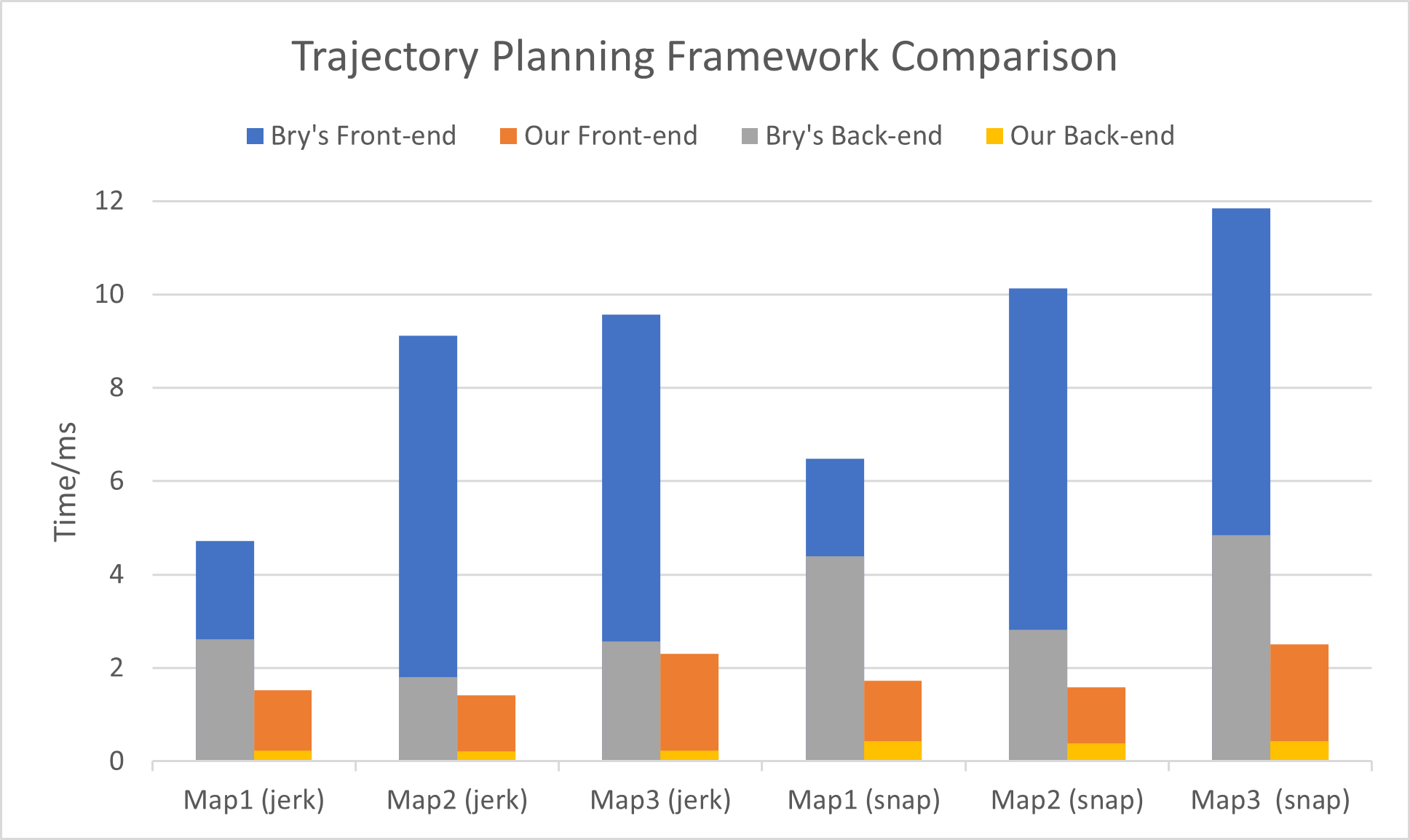}
    \caption{Trajectory planning framework comparison for minimum jerk/snap.}
    \label{fig:jerk and snap}
\end{figure}

\subsubsection{Trajectory Planning Framework Runtime Comparison}

To better compare different hierarchical trajectory planning frameworks, we have re-implemented three methods for polynomial trajectory generation. For Mellinger's method~\cite{mellinger2011minimum}, we utilize the OOQP~\cite{gertz2003object} library to solve the standard quadratic programming problem with linear equality constraints. For Bry's method~\cite{bry2015aggressive} and Oleynikova's method~\cite{oleynikova2020open}, the $A^{-1}$ and $Q$ matrices in the original paper have sparse and block-diagonal characteristics, so both dense and sparse solvers can be used for computation. In our paper, we implement it using traditional dense matrix inversion. For Wang's method~\cite{wang2022geometrically}, we directly use the open-source code provided by the authors, which employs non-singular banded PLU factorization to directly obtain the coefficient matrix of the polynomial.

In the real application of our small-scale trajectory generation problem (approximately 15 polynomial trajectory segments), Fig.~\ref{fig:jerk and snap} and Tab.~\ref{tab:table2} demonstrate the performance of our framework in three scenarios. The front end of our framework quickly generates high-quality waypoints and initial paths. Then, a polynomial trajectory that satisfies the constraints is quickly generated by combining the optimal condition and using the corresponding customized PLU decomposition. SOTA methods~\cite{bry2015aggressive}~\cite{oleynikova2020open} in trajectory planning have already shown superior performance on both quadrotors and fixed-wing drones. However, their front end uses uniform RRT* and contains many invalid samples, and their back end includes dense matrix factorization and inversion. Therefore, our method is more than five times faster compared with them, thus demonstrating significantly improved practical performance.


\subsection{Real-world Experiment}


\begin{figure*}[htb]
    \centering
    \begin{subfigure}{0.48\textwidth}
        \includegraphics[width=\textwidth]{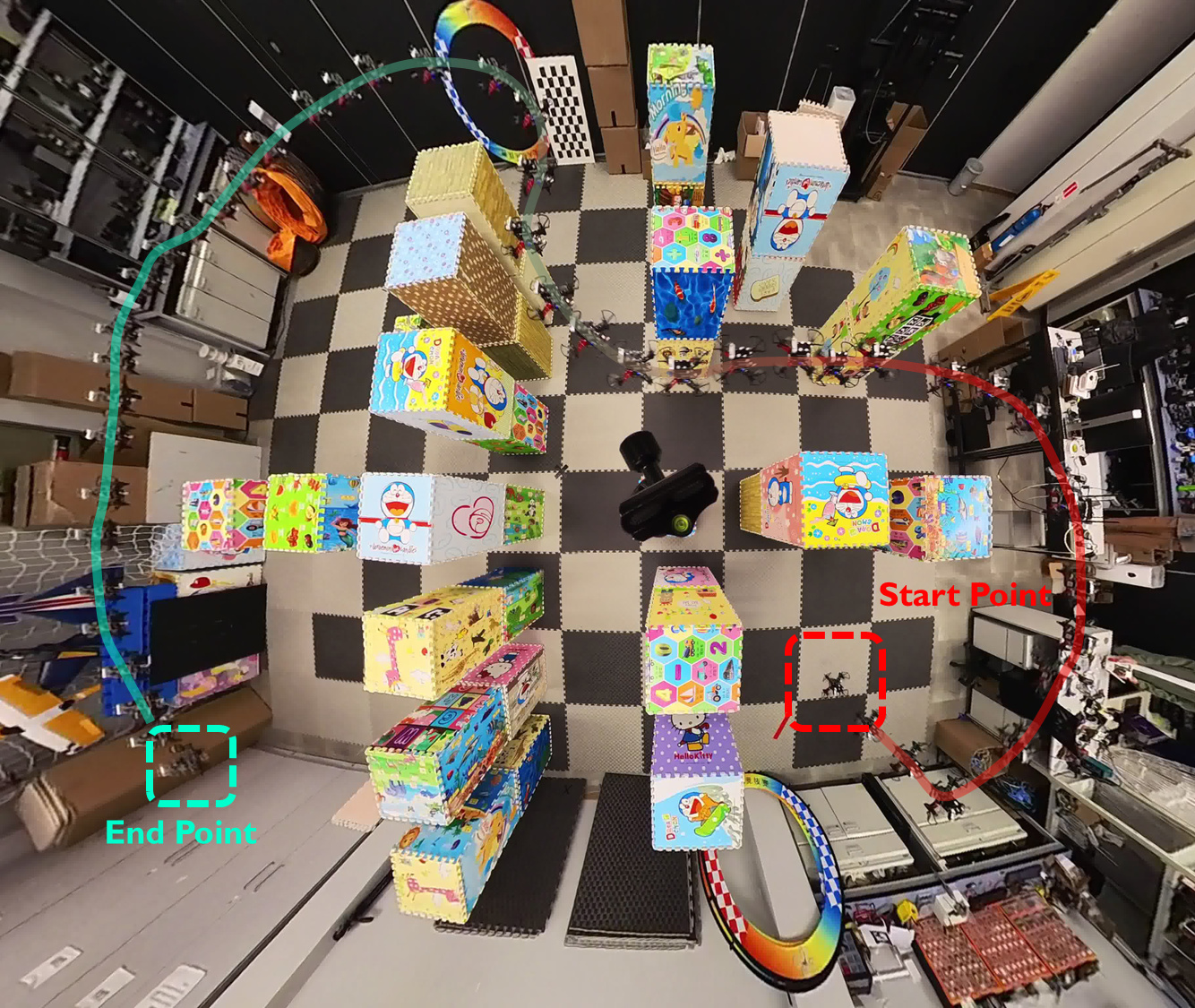}
        \caption{Real-world map1.}
        \label{fig:trajectory_a}
    \end{subfigure}
    \hspace{0.01\textwidth} 
    \begin{subfigure}{0.48\textwidth}
        \includegraphics[width=\textwidth]{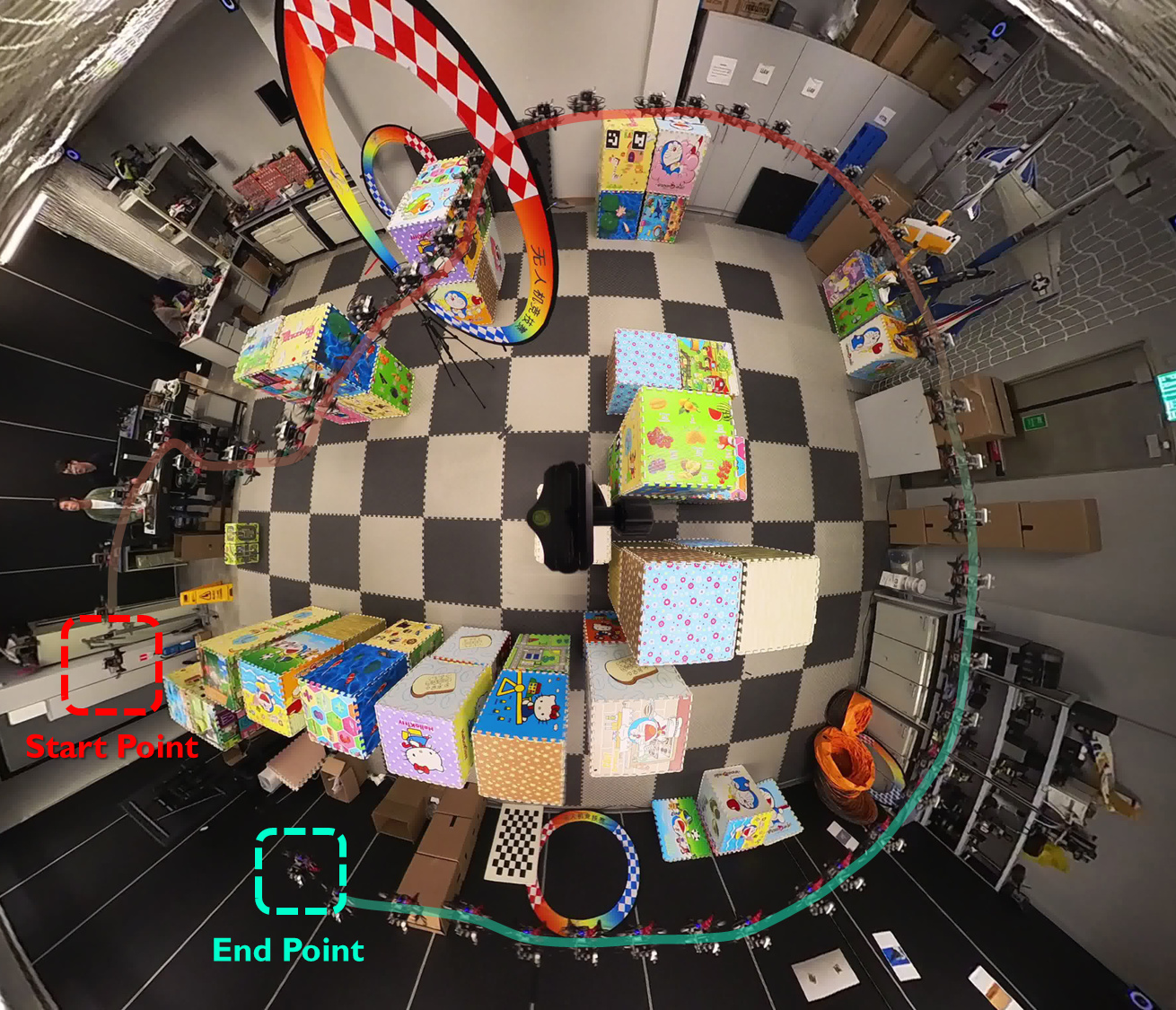}
        \caption{Real-world map2.}
        \label{fig:trajectory_b}
    \end{subfigure}
    \caption{Top view snapshot of real-world flight in two different 3D cluttered environments. The red box represents the starting point and the green box represents the end point. Each frame of flight video is combined together to form the entire UAV trajectory.}
    \label{fig:trajectory-top}
\end{figure*}

We conduct real flight tests in two cluttered and challenging indoor environments, as shown in Fig.~\ref{fig:trajectory-top}. The experimental site is an area with a size of $10m\times 10m\times 5m$. In the experimental environment, there are $15-20$ cuboid obstacles with a size of $0.6m\times 1.2m$ and a circle with a diameter of $1m$. 
The start point of trajectory planning is marked in pink color, the end point is in steel-blue color, and the actual flight trajectory is drawn with a translucent line in Fig.~\ref{fig:trajectory-top}. The average flight speeds in the two environments are $1.73km/h$ and $1.45km/h$, and the maximum flight accelerations are $0.2m/s^2$ and $0.15m/s^2$, respectively. Experimental results in the real world verify that our method can quickly find initial path solutions and generate final trajectories in 3D-cluttered environments with dead zones. The trajectory tracking mean squared error (MSE) of the UAV is shown in Tab.~\ref{tab:Tracking Error}, and the results show that our system can follow the generated trajectory with high accuracy.

\begin{table}[H]
    \caption{Trajectory Tracking Error (MSE) Analysis}
    \label{table_example}
    \begin{center}
    \resizebox{.5\textwidth}{!}{
    	\renewcommand\arraystretch{1.5}
        \begin{tabular}{c|cccccc}
        \bottomrule \hline
            Env. & Pos. Err. $x$ & Pos. Err. $y$ & Pos. Err. $z$ & Vel. Err. $x$ & Vel. Err. $y$ & Vel. Err. $z$ \\
        \hline
            Map1 & 0.0513 & 0.0919 & 0.0003 & 0.0222 & 0.0285 & 0.0006\\
            Map2 & 0.0716 & 0.1089 & 0.0004 & 0.0237 & 0.0115 & 0.0008\\
        \hline \toprule 
        \end{tabular}
    }
    \end{center}
    \label{tab:Tracking Error}
\end{table}




\section{CONCLUSION}
 
In this paper, we present a novel trajectory planning framework for quadrotor UAV autonomous flight. First, we propose a conditional generative adversarial network to generate heuristic promising regions, enabling RRT* to perform biased sampling and rapidly obtain high-quality initial paths. Second, based on the waypoints generated by the front end and the time intervals obtained from the time allocator, the back-end trajectory optimization part leverages optimality conditions to directly generate the optimal polynomial trajectory. A large number of simulations and real-world experiments demonstrate that our method is computationally much faster in complex environments compared to some SOTA approaches.

\section*{ACKNOWLEDGMENT}

The authors would like to appreciate the open-source code and hardware materials provided by HKUST Aerial Robotics Group and ZJU FAST Lab. We also appreciate the experiment site of HKUST Cheng Kar-Shun Robotics Institute.

\bibliographystyle{ieeetr}
\bibliography{main}

\end{document}